\newif\ifWithRevisions
\newif\ifWithMinorRevisions
\documentclass[letterpaper,twocolumn,10pt]{article}
\usepackage{usenix-2020-09}

\usepackage{amsmath,amsfonts,amssymb,amsthm}
\usepackage{fullpage}
\usepackage{hyperref}
\usepackage{pdfsync}
\usepackage{microtype}
\usepackage{epsfig}
\usepackage{xcolor}
\usepackage{epstopdf} 
\usepackage{graphics,graphicx,color}
\usepackage{multirow}
\usepackage{mathrsfs}
\usepackage{enumerate}
\usepackage{indentfirst}
\usepackage{subfig}
\usepackage{float}
\usepackage{hyperref}
\usepackage{tabulary}
\usepackage{thm-restate}
\usepackage{cleveref}
\usepackage{booktabs, siunitx,caption}
\usepackage{xspace}
\usepackage[normalem]{ulem}
\definecolor{DarkGreen}{rgb}{0.1,0.5,0.1}

\ifWithMinorRevisions
\WithRevisionsfalse
\newcommand{\mpcrev}[1]{\textcolor{blue}{#1}}
\newcommand{\auxrev}[1]{\textcolor{purple}{#1}}
\newcommand{\randrev}[1]{\textcolor{orange}{#1}}
\else
\newcommand{\mpcrev}[1]{{#1}}
\newcommand{\auxrev}[1]{{#1}}
\newcommand{\randrev}[1]{{#1}}
\fi

\newcommand{\attack}{\mathrm{attack}}
\newcommand{\victim}{\mathsf{honest}}
\newcommand{\adv}{\mathsf{adv}}
\newcommand{\aux}{\mathsf{aux}}
\newcommand{\shadow}{\mathsf{shadow}}
\newcommand{\subX}{X'}
\newcommand{\SensA}{A}
\newcommand{\SynthA}{A}
\newcommand{\RedactT}{\texttt{R}}

\newcommand{\victimW}{honest}

\newcommand{\squish}{
  \setlength{\topsep}{0pt}
  \setlength{\itemsep}{0ex}
  \setlength{\parskip}{0pt}
}

\newcommand{\etal}{\emph{et al.}\xspace}

\begin{document}
%

\ifWithRevisions
\newcommand{\removed}[1]{\textcolor{DarkGreen}{\sout{#1}}}
\newcommand{\added}[1]{\textcolor{blue}{#1}}

  \begin{strip}
\begin{center}\textbf{\Large{USENIX Security'21 Fall Major Revision \# 474  \\Criteria and Changes}}\end{center}
    \end{strip}

 \section*{Major Revision Criteria (Verbatim)}
 
 The reviewers found merit in the problem the authors are seeking to solve, but the paper does not clearly motivate the problem (relative to other attacks) and prove how it generalizes. As a result, we think the paper needs to make some concrete changes to warrant publication:

\begin{itemize}
    \item[-] Define and use a concrete use-case (it is not sufficient to list 3-4 generic scenarios: choose a concrete use-case and follow it through, showing that the kind of data correlation you consider actually exists in those cases. I don't mind if you then say that this generalizes to other scenarios, but please take a concrete example first.)

    \item[-] Describe the exact dependencies between the data sets (even implicit or partial ones) and argue why the attacks work (give a theoretical basis). Describe any limitations to your attacks (for instance, you say that finding individual records is out of scope -- but is it even possible with your attacks?)

    \item[-] Describe D\_aux concretely for your chosen use-case, and argue that it is reasonable knowledge for an attacker to have.

    \item[-] Perform requested changes from all reviews:
    \end{itemize}
 \noindent    \textbf{Review \#474A}
    \begin{itemize}
    \item[-] give a concrete application, with concrete types of data and attributes
    \item[-]  provide a better understanding of the limitations of your attack (with respect to the chosen assumption on the threat model) and its potential ramifications (including whether some of that data might be more correlated than it originally appears?)
\end{itemize}

 \noindent   \textbf{Review \#474B}
    \begin{itemize}
    \item[-]    Change abstract to make it clear that leakage was only found to occur when correlation is present in at least one of the other attributes or the target variable
      \item[-]     Provide data or an explanation that shows why A was categorized as correlated or not correlated with X and Y
     \item[-]      Provide data showing the "utility increase ranging from 0.5\% to 4.5\%"
      \item[-]     Support, rephrase, or remove the statement that "as we showed earlier, in realistic datasets, most of the variables are highly correlated with the sensitive variable and hence exhibit better attack accuracy"
       \item[-]    Support or remove the statement that the attack accuracy increases with a higher correlation between X' and A in the real datasets
      \item[-]     Support or remove the statement that "If the predicted split is still dominated by the same class as in the attack on the original model, it suggests that the dominant value in D\_honest2 is the same as in D\_honest1." If removed, the corresponding discussion should be reflect this change.
       \item[-]    Further support or remove the statement that Figure 3 "shows that increasing the number of queries increases the attack accuracy". Ideally this should be shown through more plotted data points in figure 3
       \item[-]    Remove or rephrase the observation that "for the 0:100 split, as few as 200 queries are sufficient to get an attack accuracy above 75
       \item[-]    Provide results for more splits in Figures 4 and 5 that represent the full range of ratios (40:60, 50:50, 60:40, 70:30, 80:20, 90:10, 100:0)
       \item[-]    Provide some discussion on why "For output class 2, attack accuracy for a model trained using the sensitive attribute (Figure 4) mostly does not change and drops less gradually than other classes in Figure 5."
       \item[-]    Support, rephrase, or remove the statement that group differential privacy would lead to high utility sacrifices
       \item[-]    Rephrase statements throughout the paper that mention that imply that an adversary can learn population statistics about another party's data in generic multi-party scenarios to make it clear that this is not the case given multiple benign parties. Instead the adversary can learn population statistics about the parties' data collectively.
\end{itemize}

 \noindent   \textbf{Review \#474D}
  
I appreciate the clarifications provided by the authors, especially on the evaluation setup and the use of $D_\aux$. Here are my feedbacks based on the rebuttal.
    \begin{itemize}
           \item[-]      Secure multi-party machine learning: I understand there is a limitation on how many words can be included in the rebuttal so please be more specific what "other information" is. In my understanding, "other than output" is too broad and too vague and cannot be proven.

          \item[-]    Evaluation setup: please clarify the split up of $D_\adv$ and $D_\victim$ earlier in the section.

         \item[-]     I still have reservations on whether the selected sensitive properties in the evaluation is truly independent of the output.
\end{itemize}
 
 \clearpage
  \section*{Changes wrt Revision Criteria}
 
{We thank the reviewers for their feedback. Below we list each of the requested changes (copied from above) and how we addressed them in this submission.}
 We also append a PDF diff between this and the original version of the paper, highlighting the text that was \added{added} and \removed{removed}, accordingly.
 \subsection*{Major Requested Changes (MR)}


\noindent{\textbf{[MR1]} \emph{Define and use a concrete use-case (it is not sufficient to list 3-4 generic scenarios: choose a concrete use-case and follow it through, showing that the kind of data correlation you consider actually exists in those cases. I don't mind if you then say that this generalizes to other scenarios, but please take a concrete example first.)}

\added{We added a use case and implication of our attacks to it in the introduction (Section~\ref{cmt:usecase})
and indicated concrete attributes and correlations that this use case may exhibit (Sections~\ref{cmt:usecase} and~\ref{cmt:correlations}).}

\vspace{5pt}
\noindent\textbf{[MR2]} \emph{Describe the exact dependencies between the data sets (even implicit or partial ones) and argue why the attacks work (give a theoretical basis). Describe any limitations to your attacks (for instance, you say that finding individual records is out of scope -- but is it even possible with your attacks?)}

\added{We give the exact correlation factors in Appendix~\ref{sec:correlation} for all datasets. We added a paragraph to provide a theoretical basis on why the attack works in Section~\ref{sec:datamodel}.}

\added{We added a paragraph on Scope in Section~\ref{sec:scope} to address limitations of our attack.}

\vspace{5pt}
\noindent\textbf{[MR3]} \emph{Describe $D_\aux$ concretely for your chosen use-case, and argue that it is reasonable knowledge for an attacker to have.}

\added{We describe approaches for generating $D_\aux$ and how they can be applied in our use case (Section~\ref{cmt:daux}).}

\vspace{5pt}
\noindent\textbf{[MR4]} \emph{Perform requested changes from all reviews.}

\added{Please see the next subsections where we address changes requested by individual reviewers.}

\subsection*{Requested Changes (Reviewer \#474A)}
\begin{enumerate}
\item \emph{give a concrete application, with concrete types of data and attributes}

\added{Please see [MR1] above.}

\item \emph{provide a better understanding of the limitations of your attack (with respect to the chosen assumption on the threat model) and its potential ramifications (including whether some of that data might be more correlated than it originally appears?)}

\added{Addressed in Scope paragraph of Section~\ref{sec:scope} and Section~\ref{sec:correlationtheory} where we expanded discussion on correlations
between attributes.}

\end{enumerate}

\subsection*{Requested Changes (Reviewer \#474B)}

\begin{enumerate}

\item \emph{Change abstract to make it clear that leakage was only found to occur when correlation is present in at least one of the other attributes or the target variable} 

\added{Addressed in the abstract.}
\item \emph{Provide data or an explanation that shows why A was categorized as correlated or not correlated with X and Y}

\added{Addressed in Appendix~\ref{sec:correlation} where we provided correlation measurements.}
  \item \emph{Provide data showing the ``utility increase ranging from 0.5\% to 4.5\%''}
  
  \added{Addressed by adding accuracy results in Table~\ref{tbl:utility} in Appendix.}
  \item \emph{Support, rephrase, or remove the statement that ``as we showed earlier, in realistic datasets, most of the variables are highly correlated with the sensitive variable and hence exhibit better attack accuracy''}
  
  \added{We removed the statement from Synthetic Data paragraph in~Section~\ref{sec:mps}. Please also see the next revision address.}
  \item \emph{Support or remove the statement that the attack accuracy increases with a higher correlation between X' and A in the real datasets}
  
  \added{We rerun the attack and updated the procedure for measuring correlations between binary categorical and
numerical variable attributes. Table~\ref{tbl:realdata-multi} shows the updated results that support the statement in~Section~\ref{sec:mps}.}

  \item \emph{
     Support or remove the statement that ``If the predicted split is still dominated by the same class as in the attack on the original model, it suggests that the dominant value in $D_\mathsf{honest2}$ is the same as in $D_\mathsf{honest1}$.'' If removed, the corresponding discussion should be reflect this change.}

     \added{Addressed in Section~\ref{sec:cmtrevB6} by rephrasing the attack strategy.}
  
   \item \emph{  Further support or remove the statement that Figure 3 ``shows that increasing the number of queries increases the attack accuracy''. Ideally this should be shown through more plotted data points in Figure~3} 
   
   \added{Addressed in Section~\ref{sec:param} by removing the statement.}
      \item \emph{  Remove or rephrase the observation that ``for the 0:100 split, as few as 200 queries are sufficient to get an attack accuracy above 75\%'' so that it is consistent with the other provided data points}
        
        \added{Addressed in Section~\ref{sec:param} by rephrasing the statement.} 

       \item \emph{ Provide results for more splits in Figures 4 and 5 that represent the full range of ratios (40:60, 50:50, 60:40, 70:30, 80:20, 90:10, 100:0)}  
       
       \added{Addressed in Section~\ref{sec:param} and Figure~\ref{fig:graph_removetrue} by adding data points for all splits.}
      \item \emph{  Provide some discussion on why ``For output class 2, attack accuracy for a model trained using the sensitive attribute (Figure 4) mostly does not change and drops less gradually than other classes in Figure 5'}' 
      \added{Addressed in Section~\ref{sec:param} by providing a potential explanation.}
       \item \emph{ Support, rephrase, or remove the statement that group differential privacy would lead to high utility sacrifices}
       
       \added{Addressed in Section~\ref{cmt:revB11} by rephrasing the statement.}
    
    \item\emph{Rephrase statements throughout the paper that mention that imply that an adversary can learn population statistics about another party's data in generic multi-party scenarios to make it clear that this is not the case given multiple benign parties. Instead the adversary can learn population statistics about the parties' data collectively.}
    
    \added{Addressed throughout the paper, including Sections~\ref{sec:mpcml} and~\ref{sec:attack} where $D_\victim$ is defined as a dataset that corresponds to the data of the
(victim) parties whose data is not known to the attacker and the property that the attacker is trying to infer is defined wrt $D_\victim$.}
    \end{enumerate}

 \subsection*{Requested Changes (Reviewer \#474D)}}
 \begin{enumerate}
\item \emph{Secure multi-party machine learning: I understand there is a limitation on how many words can be included in the rebuttal so please be more specific what "other information" is. In my understanding, ``other than output'' is too broad and too vague and cannot be proven.}
    
    \added{We added a paragraph on secure multi-party computation and its guarantees in Section~\ref{sec:smpc}.}
\item \emph{Evaluation setup: please clarify the split up of $D_\adv$ and $D_\victim$ earlier in the section.}

   \added{We added a paragraph with details in the Dataset split in Section~\ref{sec:dataatt_split}.}

\item \emph{I still have reservations on whether the selected sensitive properties in the evaluation is truly independent of the output.}

\added{For experiments with a synthetic sensitive attribute, we can control how synthetic attribute influences other attributes. Please see~Section~\ref{sec:synthgen} (Synthetic Data) for details.
For real datasets, it is indeed impossible to say that variables are independent. To this end, we added an explanation of how dependency was determined for each dataset based on correlation factors in Appendix~\ref{sec:correlation}.}
    \end{enumerate}

  \subsection*{Changes to the paper}
  In addition to addressing all the revisions as specified above, we summarize main changes to the paper below:
   \begin{itemize}
\item We added Appendix~\ref{sec:correlation} where we specify correlation coefficients and how dependency between attributes in real datasets was determined;
\item We now use point-biserial correlation coefficient to determine correlation between binary categorical and numerical variables --- this resulted in changes to \# $X'$ in Table~\ref{tbl:realdata-multi};
\item We have rerun our experiments with query set $D_\adv$ based on random and not artificially constructed records (inline with attack strategy in Section~\ref{sec:attack}) --- this resulted in changes to accuracies in Table~\ref{tbl:realdata-multi};
\item We have added experiments where smaller size of~$D_\aux$ was used, Table~\ref{tbl:realdata-smallaux} in Appendix.
\item We have updated Figures~\ref{fig:graph_removetrue} and ~\ref{fig:graph_removefalse} to include data for all splits. We moved the Figure~\ref{fig:graph_removefalse} to Appendix.
\item We moved white-box attack experiments to Appendix~\ref{sec:whiteboxapp}.
\end{itemize}

\else
\newcommand{\added}[1]{{#1}}
\newcommand{\removed}[1]{}
\fi

\pagestyle{empty}

\ifWithMinorRevisions

  \begin{strip}
\begin{center}\textbf{\Large{USENIX Security'21 Winter Minor Revision \# 300}}\end{center}
    \end{strip}

Reviewers' ``comments to author'' are pasted below.
The next page lists how the comments were addressed.

\section*{ Comments for author}

\vspace{5pt}
 \noindent    \textbf{Review \#300A}
 
I am satisfied that my personal concerns were answered. Thanks for the efforts you have put into the revision.

\vspace{5pt}
 \noindent    \textbf{Review \#300B}

Thank you to the authors for updating the paper as requested! I think this paper is much better motivated than in the previous round. The authors present a more compelling case for why this type of attack is relevant; I thought the banking example was useful, and the use of concrete examples to motivate different data correlation models significantly clarified that section.

I also thought the experimental results were more clearly explained (perhaps this is partly due to the clarified exposition in the front half of the paper).

I have only one minor issue: I didn't quite understand the role of MPC in this work. Perhaps the authors are proposing that it is used to train the model in a black-box way? That is reasonable but not explicitly spelled out.

\vspace{5pt}
 \noindent    \textbf{Review \#300C}

The revision does a good job at incorporating feedback from last
round of reviews. The revised Section 3 does a much better job on
explaining the problem setup and why the information leakage can
happen. The evaluation setup is also more clear. The authors also
include better comparison with related work and how some existing
defense mechanism can help eliminate/reduce the unintended leak.

\vspace{5pt}
 \noindent    \textbf{Review \#300D}

The authors have clarified the model, limitations, and the properties of the dataset used in the attack, and the revised version is more clear to understand its problem scope and the claim that the paper makes.

But I think that the practical effect of the attack is quite limited. This is because the attack would work well in the case that the attacker may obtain the dataset with similar property distribution, and also the accuracy is quite low. Although the attack accuracy is better than the random guess, that is, around 0.6 > 0.5 (random guess) in Table 5 and around 0.5 > 0.2 (random guess), however, I am curious about the practical implication of such numbers. If this is an attack to cryptographically keys or other data that can enhance the attack for the other methods, such as, decreasing the time for brute-force attacks, then having a slightly higher attack accuracy is very meaningful to the attackers, however, in a case that the attacker would like to know the single answer for the sensitive property, I think that the accuracy ranged from 0.4$\sim$0.6 still gives uncertainty to the attackers.

I believe the authors finding and the problem is interesting, but the practical impact of this attack is still in question.

\clearpage

 \section*{Revisions}

{We thank the reviewers for their feedback and kind comments. Below we list addressed reviewers' suggestions
and attach the paper with the corresponding changes.}


\vspace{9pt}
\noindent{\textbf{[\#300B]} \emph{I have only one minor issue: I didn't quite understand the role of MPC in this work. Perhaps the authors are proposing that it is used to train the model in a black-box way? That is reasonable but not explicitly spelled out.}

\added{Addressed in \mpcrev{blue colour} in the paper below in Sections~\ref{sec:prelims},~\ref{sec:attack} and~\ref{sec:eva}.}

\vspace{9pt}
\noindent\textbf{[\#300D-1]} \emph{But I think that the practical effect of the attack is quite limited. This is because the attack would work well in the case that the attacker may obtain the dataset with similar property distribution}

\added{Addressed in \auxrev{purple colour} in the paper below in Sections~\ref{sec:prelims} and~\ref{sec:attack}.}

\vspace{9pt}
\noindent\textbf{[\#300D-2]} \emph{Although the attack accuracy is better than the random guess, that is, around 0.6 > 0.5 (random guess) in Table 5 and around 0.5 > 0.2 (random guess), however, I am curious about the practical implication of such numbers. If this is an attack to cryptographically keys or other data that can enhance the attack for the other methods, such as, decreasing the time for brute-force attacks, then having a slightly higher attack accuracy is very meaningful to the attackers, however, in a case that the attacker would like to know the single answer for the sensitive property, I think that the accuracy ranged from 0.4$\sim$0.6 still gives uncertainty to the attackers.}

\added{Addressed in \randrev{orange colour} in the paper below in Section~\ref{sec:experiments}.}

\added{We note that as described in~Section~\ref{sec:mps}
attack accuracy depends on the correlation between~$X$ and~$A$ for the $X\sim A, Y\bot A$ setting.
In \textbf{Synthetic Data} experiments in~Table~\ref{tbl:synthetic-multi}}, the correlation is low for the $X \sim A, Y \bot A$ setting
and, hence, attack accuracy is low. As we increase the signal, i.e., the $X \sim A, Y \bot A$ (\RedactT{}) setting, the attack accuracy becomes higher.
Finally, in practice for \textbf{Real Data} experiments in Table~\ref{tbl:realdata-multi}, we observe strong correlation and, hence, high accuracy.}

\fi

\title{Leakage of Dataset Properties in Multi-Party Machine Learning}
\author{
{\rm Wanrong Zhang\footnotemark[4]}\\
Georgia Institute of Technology
\and
{\rm Shruti Tople}\\
Microsoft Research
 \and
{\rm Olga Ohrimenko\footnotemark[4]}\\
The University of Melbourne
} 
\renewcommand{\thefootnote}{\fnsymbol{footnote}}
\maketitle
\footnotetext[4]{Work done in part while at Microsoft.}
\renewcommand{\thefootnote}{\arabic{footnote}}
\begin{abstract}
Secure multi-party machine learning allows several parties to build a model on their pooled data to increase utility while not explicitly sharing data with each other. 
We show that such multi-party computation can cause leakage of global dataset properties between the parties
even when parties obtain only black-box access to the final model.
In particular, a ``curious'' party can infer the distribution of sensitive attributes in other parties' data with high accuracy.
This raises concerns regarding the confidentiality of properties pertaining to the whole dataset as opposed
to individual data records.
We show that our attack can leak {\em population-level} properties in datasets of different types, including tabular, text, and graph data.
To understand and measure the source of leakage, we consider several models of correlation between a sensitive attribute and the rest of the data.
Using multiple machine learning models, we show that leakage occurs even if the sensitive attribute is {\em not} included in the training data and has a low correlation with other attributes \removed{and}\added{or} the target variable.

\end{abstract}


\section{Introduction}

Modern machine learning models have been shown to memorize information
about their training data, leading to privacy concerns regarding their use and release in practice.
Leakage of sensitive information about the data has been shown via membership attacks~\cite{DBLP:conf/sp/ShokriSSS17,Salem:NDSS19},
attribute inference attacks~\cite{Song2020Overlearning,DBLP:conf/ccs/FredriksonJR15},
extraction of text~\cite{DBLP:journals/corr/abs-1802-08232}
and data used in model updates~\cite{DBLP:conf/uss/0001B0F020,brockschmidt2019analyzing}.
These attacks focus on leakage of information
about an individual record in the training data,
with several recent exceptions~\cite{10.1145/3243734.3243834,DBLP:journals/corr/abs-1805-04049} pointing out
that leakage of global properties about a dataset
can also lead to confidentiality and privacy breaches.

In this paper, we study the problem of leakage of dataset properties at the {\em population-level}. Attacks on leakage of global properties about the data are concerned with learning information about the \emph{data owner} as opposed to individuals whose privacy may be violated via membership or attribute inference attacks. The global properties of a dataset are confidential
when they are related to the proprietary information or IP that the data contains, and its owner is not willing to share.
\added{As an example,} consider the advantage one can gain \removed{by}\added{from} learning demographic information of customers or sales distribution across competitor's products. \removed{Concerns on whether machine learning models trained on the data can leak global properties about a sensitive attribute (e.g., sales numbers or gender) arise in several settings, including a single party releasing a trained model and multi-party machine learning.}

Our primary focus is on inferring dataset properties in the \emph{centralized multi-party machine learning setting}. \added{This setting allows multiple parties to increase utility of their data since the model they obtain is trained on a larger data sample
than available to them individually.
Benefits of computing on combined data have been identified in multiple sectors
including drug discovery,
health services, manufacturing and finance.
\label{cmt:usecase}
For example, anti-money laundering served as a use case for secure data sharing and computation
during the TechSprint organized by the Financial Conduct Authority, UK in 2019~\cite{fca-techsprint}.
A potential machine learning task in this setting is to create a system that identifies
a suspicious activity based on financial transactions and demographic information
of an entity (e.g., a bank customer). Since multiple financial institutions have separate views of the activities, such
a system can be used to detect common patterns.
}

\added{Deployments and availability of secure computation methods~\cite{acc,sealcrypto,ibmsgx,crypten2020} can enable
multi-party machine-learning by alleviating immediate privacy concerns of the parties.
In particular, secure multi-party machine learning provides parties with a \emph{black-box} access to a model trained on their pooled data
without requiring the parties to share plaintext data with each other.}
\added{Unfortunately, as we show in this paper, this is insufficient to address
all privacy implications of collaborative machine learning.
In particular, we demonstrate that} \emph{global properties} about \removed{of the parties'}
\added{one party's} sensitive attributes can be inferred by the second party, even
\added{when only black-box access to the model is available.}
\added{Consider implications of our attacks in the use case above.
\label{cmt:usecaseattack}
An attacker party (e.g., one of the banks) can learn distribution
of demographic features pertaining to the customer population in the other bank
(e.g., whether the other bank has more female than other customers or what percentage of customers
has income over a certain threshold)
that it can use in the future when developing a marketing campaign to attract new customers.}

\added{
Analysis of our attacks shows that leakage of population-level properties is possible even
in cases where sensitive attribute is irrelevant to the task, i.e., it has $\approx0$ correlation with the task in hand.}
Though removing sensitive attributes may seem like a viable solution, it is not provably secure due to correlations
that are present in the data.
Indeed, we show that in many cases, information is still leaked regardless of whether training data contained the sensitive attribute or not.
\added{We argue that this is possible due to correlation between sensitive attributes and
other attributes that exists in the data.
For example, datasets we use indicate that there is correlation between sets of attributes including gender, occupation and working hours per week,
as well as income, occupation and age. Such customer attributes are often recorded by financial institutions,
as a result indicating potential leakage if institutions were to collaborate towards detection of financial crime as described above.}
\removed{Consider the following example. In the product recommendation setting, data of two retailers is joined to improve utility for both when training a model for a wide variety of products.
We show that one of the parties (an attacker)
can learn product distribution in another party's data,
as a result, revealing which product type sells best for the other retailer.
In another example, two hospitals may wish to train a model securely
on their pooled data to predict the number of days a patient will stay in a hospital.
We show that given a model trained for this task,
one of the hospitals can learn whether the other hospital has more female than male patients.
In fact, we show that this leakage is possible even
though gender attribute is irrelevant to the task, i.e., it has $\approx0$ correlation with the number of days patients stay in a hospital.}

\begin{table*}[t]
\centering
\begin{tabular}{l|l|l|l|l}
 & \bf{Attacker's knowledge}  & \bf{Single-party} & \bf{Multi-party} & \bf{Datasets}\\
 \hline
Melis~\textit{et al.}~\cite{DBLP:journals/corr/abs-1805-04049} & training gradients & & $\checkmark$  & tabular, text, images\\
Ganju~\textit{et al.}~\cite{10.1145/3243734.3243834} & model parameters (white-box) & $\checkmark$ & & tabular, images \\
 Ateniese~\textit{et al.}~\cite{10.1504/IJSN.2015.071829} & model parameters (white-box) & $\checkmark$ & & tabular, speech \\
 \hline
This work  & model predictions (black-box)& $\checkmark$ & $\checkmark$ &tabular, text, graphs 
\end{tabular}
	\caption{Comparison of attacks on leakage of dataset properties.}
	\label{tbl:attackcmp}
\end{table*}

\vspace{-5pt}
 \paragraph{Threat model.}

We consider the setting where the model is securely trained on the joined data of the honest party and of
an \emph{honest-but-curious} party.
Honest-but-curious adversary considers a realistic setting where the malicious party (1) will not alter
its own data --- if it does, the model may not perform well and, if detected, could undermine the trust
from the other party in the partnership ---
and (2) will not change the machine learning code --- both parties may wish to observe
the code to be run on the data to ensure its quality and security.

The attacker is interested in learning global properties
about a sensitive attribute {at the dataset level},
that is, how values of this attribute are distributed in the other party's dataset.
It may be interested in learning which attribute value is dominant
(e.g., whether there are more females)
or what the precise ratio of attribute values is (e.g.,
90\% \added{fe}males vs.\added{~70}\% females).
%

\paragraph{Attack technique.}
We show that dataset property can be leaked merely from the black-box access
to the model. In particular, the attacker does not require access to the training process of the model (e.g., via gradients~\cite{DBLP:journals/corr/abs-1805-04049})
or to model parameters (aka white-box attack~\cite{10.1145/3243734.3243834,10.1504/IJSN.2015.071829}).
Following other attacks in the space,
the attacker also uses shadow models and a meta classifier.
However,
individual predictions from the model are not sufficient to extract global information about a dataset.
To this end, we introduce an attack vector based on a set of queries
and use them in combination in order to infer a dataset property.
In contrast to previous work on property leakage, the attack
requires less information and assumptions on the attacker
(see Table~\ref{tbl:attackcmp} and Section~\ref{sec:related} for more details).

\vspace{-5pt}
\paragraph{Methodology.}
To understand what causes
information leakage about a property we consider several correlation relationships
between the sensitive attribute~$A$, the rest of the attributes~$X$,
and the target variable~$Y$ that the machine learning model aims to learn.
Surprisingly, we show that dataset-level properties about~$A$ can be leaked
in the setting where~$A$ has low or no correlation with~$Y$.
We demonstrate this
with experiments
on real data and experiments with a synthetic attribute
where we control its influence on $X$ and $Y$.
The attack persists across different model types such as logistic regression,
multi-layer perceptrons (MLPs), Long Short Term Memory networks (LSTMs), and Graphical Convolution Networks (GCNs) models
and for different dataset types such as tabular, text, and graph data. The attack is efficient as it requires 100 shadow models and fewer than $1 000$ queries.

\vspace{-5pt}
\paragraph{Machine learning settings.}
In addition to the multi-party setting,
our property leakage attack can be carried out in the following two settings.
(1) single-party setting
where an owner of a dataset releases query interface of their model;
(2) in the model update setting,
one can infer how the distribution of a sensitive
property has changed since the previous release of the model.
The second attack also applies to multi-party machine learning,
showing that the party that joins last exposes its data distribution
more than parties who were already collaborating.

\paragraph{Contributions.} Our contributions are as follows: 
\begin{itemize}
\squish
         \item {\em Problem Formulation:} We study leakage of properties about a dataset used to train a machine learning model when only black-box access to the model is available to the attacker. 
         \item {\em Attack Technique:} We propose an effective attack strategy that requires only a few hundred inference queries to the model (black-box access) and relies on a simple attack architecture that even a computationally bound attacker can use. 
         \item {\em Attack Setting:} We show that leakage of dataset properties is an issue for an owner of a dataset when the owner releases a model trained on their data (single-party setting); when the owner participates in multi-party machine learning, and when the owner contributes data to update an already trained model (e.g., either because it joins other parties or because it has acquired new data).
         \item {\em Empirical Results:}  We show that distribution of a sensitive attribute can be inferred with high accuracy for several types of datasets (tabular, text, graph) and models, even if the sensitive attribute
         is dropped from the training dataset and has low correlation with the target variable.
\end{itemize}

Finally, we note that secure multi-party \removed{machine learning}\added{computation},
based on cryptographic techniques or secure hardware,~\cite{7958569,10.1145/2508859.2516751,10.1145/3338466.3358924,10.1007/978-3-662-45472-5_12,10.1145/1150402.1150477,10.1007/978-3-642-37682-5_1,Ohrimenko2016,DBLP:journals/corr/abs-1807-06689,DBLP:journals/corr/abs-1803-05961,tramer2018slalom} guarantees that
nothing except the output of the computation is revealed to the individual parties.
However, it is not concerned with what this final output can reveal about the input data of each party.
On the other hand, defenses, such as differential privacy,
are concerned with individual record privacy and not dataset property privacy
considered in this paper. We discuss this further in~Section~\ref{sec:defences}.
In summary, we believe this work
identifies a potential gap in multi-party machine learning research in terms of techniques that parties can deploy
to protect global properties about their dataset.


\section{\added{Preliminaries}} \label{sec:prelims}

We assume that there is an underlying data distribution~$\mathcal{D}$
determined by variables $X$, $A$, $Y$ where
$X$ models a set of features, $A$ models a feature
that is deemed private (or sensitive) and $Y$ is the target variable, i.e., either a label or
a real value (e.g., if using regression models).
We consider a supervised setting 
where the goal is to train a model $f$ such that $f(X,A)$
predicts $Y$.

\added{\paragraph{Secure multi-party computation (MPC).}
\label{sec:smpc}
MPC lets parties obtain
a result of a computation on their combined datasets
without requiring them to share plaintext data
with each other or anyone else.
Methods that instantiate it
include homomorphic encryption,
secret sharing, secure hardware
and garbled circuits~\cite{DBLP:conf/eurocrypt/PassST17,10.1007/978-3-642-32009-5_38,10.1007/978-3-662-48000-7_4,10.5555/2028067.2028102,10.1145/73007.73014}.
These methods vary in terms of their
security guarantees (e.g., availability of a trusted processor vs.~non-colluding servers) and efficiency.
We abstract MPC using an ideal functionality~\cite{DBLP:conf/eurocrypt/PassST17}:
a trusted third entity accepts inputs
from the parties, computes the desired function on the combined data,
and returns the output of the computation to each party.
Security of protocols implementing this functionality is often captured
by proving the existence of a simulator that can simulate adversary's view in the protocol
based only on adversary's input and the output of the
computation. Hence, an MPC protocol guarantees that an adversarial party learns only the output of the computation
but does not learn the content of the inputs of other parties beyond what it can infer based on its own data and the output.
}
\removed{However, }
\added{Since our attacks are oblivious to the exact technique used for secure} \removed{training}\added{computation},
\added{we assume ideal MPC functionality and specify additional information available
to the adversary in the next section.}

\paragraph{Multi-party machine learning.}
\label{sec:mpcml}
Let $D_\victim$ and $D_\adv$ be the datasets corresponding to the data of the
victim parties and $D_\adv$ be the data that belongs to the parties
whose data is known to the adversary.
For simplicity, we model it using two parties $P_\victim$ and $P_\adv$
who own $D_\victim$ and $D_\adv$, respectively.
\auxrev{Both $D_\victim$} and $D_\adv$ are sampled from $\mathcal{D}$ \auxrev{but may have a different distribution of~$A$,}
conditional on some latent variable, for example, a party identifier.
\auxrev{Importantly,  distribution of $A$ in $D_\victim$ is secret and unknown to $P_\adv$.}
Parties are interested in increasing the utility of their model
through collaboration with each other.
\added{To this end, they agree on an algorithm
to train} \removed{The} \added{a machine learning} model,~$f$, \removed{is trained on}\added{using their} combined
\added{datasets} $D_\victim$ and $D_\adv$.

\mpcrev{\added{The parties use secure multi-party computation to train~$f$, as they are not willing to share it
either due to privacy concerns or regulations.}}
\removed{The trained} \added{Once the} target model \mpcrev{\added{is trained using MPC}, it} can be released to the parties
either as a white- or black-box.
In the former, $f$ is sent to the parties,
and, in the latter, the model is available to the parties through an inference interface
(e.g., the model stays encrypted at the server such that inferences are made
either using secure hardware or cryptographic techniques~\cite{10.5555/3277203.3277326}).
\mpcrev{We assume \added{that}~$f$ is trained faithfully \removed{using secure multi-party machine learning}}
and, hence, $P_\adv$ \removed{has no additional information about the content of~$D_\victim$}
\added{cannot}\removed{nor can} tamper with how $f$ is trained (e.g., this avoids attacks where a malicious algorithm can encode training data in model weights~\cite{Congzheng}).

 \removed{Secure multi-party machine learning}\mpcrev{\added{MPC} guarantees
that parties learn nothing about the computation besides the output, i.e.,
they learn no other information about each other's data besides what is revealed
from their access to~$f$.} The goal of this paper is to show that
even by having black-box access to $f$ one party can infer information about \removed{the} other \removed{party's}\added{parties'}
data.

\section{\added{Data Modeling}}
\label{sec:datamodel}
To reason about leakage of $A$'s distribution in~$\mathcal{D}$, we consider
different relationships between $X,Y,A$ based on their correlation.
We use $\sim$
to indicate that there is a {correlation} between random variables
and $\bot$ if not.
{\removed{Correlation between variables.}}
\label{sec:correlationtheory}
We consider four possible relationships between $Y$, $X$
and the sensitive attribute $A$.

\vspace{5pt}
\noindent
\added{\underline{$Y \bot A$}:}
If $Y$ is independent of $A$,
and if $f$ is faithfully modeling
the underlying distribution, $A$ should not be leaked.
That is, information about $A$ that an adversary
acquires from $f(X,A)$ and $f'(X)$ should be the same
for models $f$ and $f'$ trained to predict $Y$.
Two scenarios arise depending on whether
the rest of the features are correlated with~$A$ or not:
{$(\boldsymbol{X\bot A, Y\bot A)}$} and $(\boldsymbol{X \sim A, Y \bot A)}$.
\added{We argue that leakage in the latter case is possible due
to how machine learning models
are trained. Below we describe why it is theoretically feasible and
experimentally validate this in Section~\ref{sec:experiments}.}

\added{
A machine learning model is trying to learn the conditional probability distribution $\Pr(Y=y|X=x)$ where~$X$ are the attributes and $Y$ is the target variable.  Suppose there is a latent variable~$Z$, and the observed~$X$ is modeled by $X=h(Z, A)$ where $h$ is a function capturing the relationship between the variables. Even if the target variable $Y$ only depends on $Z$
through a random function $g$: $Y=g(Z)$, the conditional distribution $\Pr(Y=y|X=x)$ still depends on $A$.
Thus, machine learning models will capture information about $A$.
For example, consider a task of predicting education level ($Y$) based on data that contains gender ($A$) and income ($X$).
Suppose income can be modeled by a function of latent variables skill and occupation, and education level is only associated with the skill. 
Though gender is not correlated with education level ($Y\bot A$), it could be associated with occupation and thus correlated with income ($X$).}

The $(X\sim A, Y \bot A)$ scenario was also noted by Locatello~\textit{et al.}~\cite{locatello2019fairness}
when studying fair representations.
The authors indicated that even if the original data may not have a bias
(i.e., when the target variable and the protected variable are independent)
using the protected attribute in training can introduce bias.

To model  $(X \sim A, Y \bot A)$ scenario in the experiments,
we use correlation coefficients to determine the split of dataset attributes into $X$ and $A$.
To have a more controlled experiment, we
also carry out experiments where we introduce a synthetic variable and inject correlations between
it and a subset of attributes in $X$.

\vspace{5pt}
\noindent
\added{\underline{{$Y \sim A$}:}}
We also consider two cases where there is a correlation between the target variable $Y$
and the sensitive attribute~$A$:
{$(\boldsymbol{X\bot A, Y\sim A)}$} and $(\boldsymbol{X \sim A, Y \sim A)}$.
{In the setting of $(X \bot A, Y \sim A)$, attribute $A$ and a set of attributes~$X$ may be relevant in predicting $Y$,
while being uncorrelated with each other.
For example, a reaction of an individual to a new drug ($Y$)
could depend on the age and weight of an adult,
while age and weight
may be regarded as
independent between each other.}

The final setting of $(X \sim A, Y \sim A)$
is the most likely scenario
to happen in practice where the true distribution
and dependence between variables
maybe unknown.
\added{\label{cmt:correlations}For example, consider a task of predicting whether
a financial transaction by an individual is suspicious or not~($Y$)
based on customer information (e.g., occupation, age, gender) and their transaction history ($X$),
where their income is the sensitive attribute $A$.
The correlation between attributes could either belong to cases $(\boldsymbol{X \sim A, Y \bot A)}$
or to $(X \sim A, Y \sim A)$ since attributes such as occupation and age are likely to be correlated with income
(as also suggested by the correlations in the datasets we use in our experimental evaluation in~Appendix~\ref{sec:correlation}).}

\section{Threat Model and Attack}
\label{sec:attack}

The goal of the adversarial party $P_\adv$ is to learn population-level properties
about the \added{rest of the} dataset \removed{of the other party} \added{used} in the multi-party machine learning setting
\added{(e.g., in the two-party setting this corresponds to learning properties of the other party's dataset)}.
Since~$P_\adv$ is one of the parties, it has
black-box access to
the joint model~$f$ trained \mpcrev{(e.g., via MPC)} on the data of all the parties (i.e., $D_\victim$ and $D_\adv$).
Given this query interface to~$f$, the attacker wants to infer how sensitive 
attribute~$A$ is distributed in
\victimW{} \removed{party's}\added{parties'} dataset~$D_\victim$.
Throughout the paper, we use attribute and feature interchangeably.

We model dataset property leakage as follows.
Let ~$\mathbf{a_\victim}$
denote attribute values of $A$ for all records in~$D_\victim$
(for example, if the sensitive attribute is gender, then $\mathbf{a_\victim}$
is a vector of gender values of all records in $P_\victim$ data).
We define $p(\mathbf{a_\victim})$ to be the property or information about
$\mathbf{a_\victim}$ 
that the adversary is trying to infer.
For example, the property could be related to determining
whether there is a higher presence of female
patients in the dataset $D_\victim$
or learn the exact ratio of female patients.

The attacker,
besides knowing its own dataset~$D_\adv$ and having black-box access to the model $f$,
\auxrev{is assumed to have auxiliary dataset~$D_\aux$
that is distributed according to~$\mathcal{D}$.}\removed{(e.g., a publicly available dataset)}
\added{\label{cmt:daux}Similar to~\cite{DBLP:conf/sp/ShokriSSS17}, an auxiliary dataset can be generated 
either via (1) model-based synthesis approach --- feeding synthetic data to $f$ and using its output to guide the search towards
data samples on which the model returns predictions with high confidence,
(2) statistics-based synthesis that uses information about marginal distribution of the attributes, or
(3) using a (publicly available) dataset of similar distribution.
The attacker can use approach (1) by merely using $f$, while $D_\adv$
provides it with statistics for (2). The availability of a dataset that follows similar
distribution to $\mathcal{D}$ depends on the setting.
Consider the anti-money laundering use case in the introduction.
A party may have access to billions of financial transactions that it can use either for approach (2)
since record-level marginal distribution between demographic features, income, education level is likely
to be similar between the parties, or for approach (3) by dividing its dataset into $D_\aux$ and $D_\adv$.}

\begin{figure*}[t]
	\centering
	\includegraphics[scale=0.51]{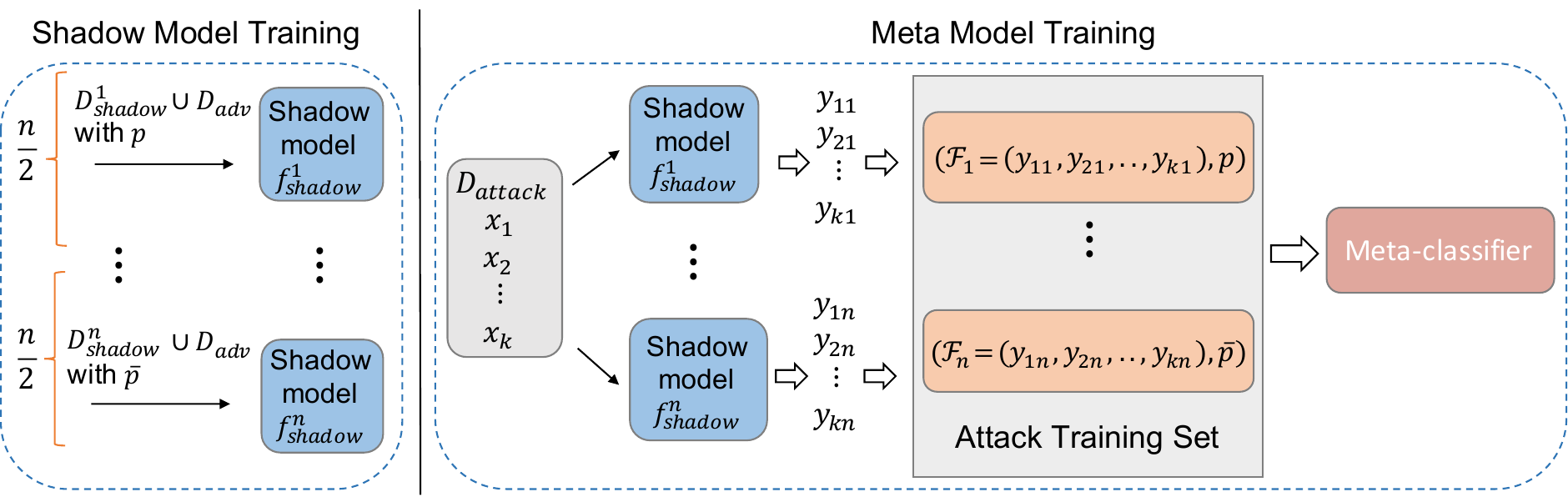}
	\caption{Attack model pipeline. Half of shadow models are trained with the property $p$ that the attacker is trying to learn and half without it.
	Each shadow model $f^i_\shadow$ is queried
	on a dataset $D_\attack$. Output probability vectors are concatenated to form a vector $\mathcal{F}_i$.
	Finally, the meta-classifier is trained on feature-label tuples of the form~$\{(\mathcal{F}_i,p_i)\}_i$.}
	\label{fig:attack}
\end{figure*}

\begin{figure*}[t]
	\centering
	\includegraphics[scale=0.51]{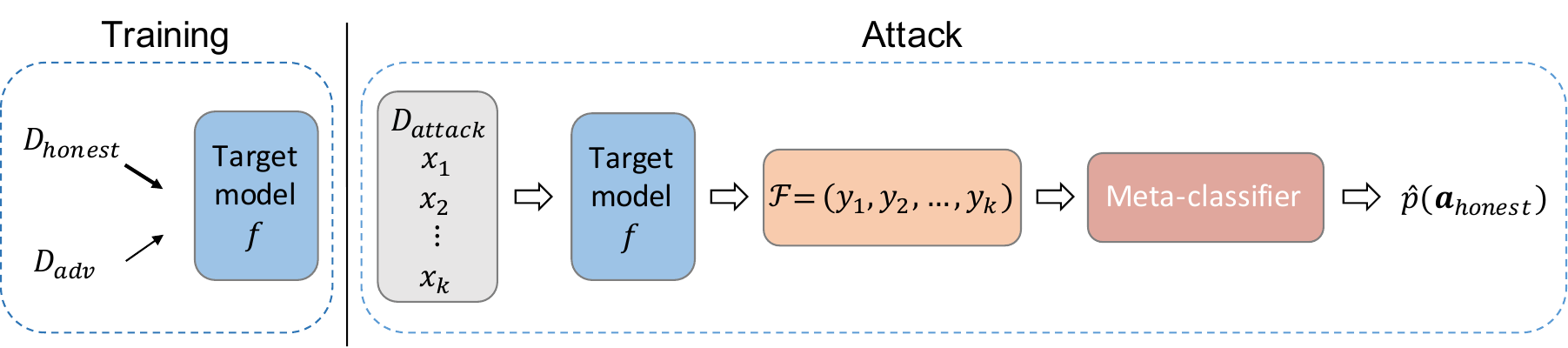}
	\caption{Execution of the attack on the target model to learn the prediction of the property $p(\mathbf{a}_\victim)$ in $D_\victim$, $\hat{p}$.}
	\label{fig:actualattack}
\end{figure*}

\removed{Though our}\added{The} attack \added{follows}\removed{is based on} the shadow model training approach~\cite{10.1504/IJSN.2015.071829,DBLP:conf/sp/ShokriSSS17}.
\added{However,}
we modify the attack vector to measure the signal
about the distribution of a sensitive attribute in a whole dataset.
Our attack strategy is described below;
Figure~\ref{fig:attack} shows graphical representation of how the attack model is trained
and Figure~\ref{fig:actualattack} shows the execution of an attack on target model~$f$.

\removed{Our approach is based on the observation}
\added{We make an observation}
that to infer global properties about training data,
the attacker needs to combine information from
multiple inferences made by~$f$.
To this end, the attacker measures
how~$f$ performs
on a sequence of $k$ records, called $D_\attack$, as opposed to a single record used in work on attribute and membership inference.
We obtain the ``attack feature'' sequence $\mathcal{F}$ by setting it to the
posterior probability vector across classes
returned by $f$ on $D_\attack$.
Hence, if $f$ is a classification model over $l$ classes
$\mathcal{F}$ consists of $k \times l$ values.
In the experiments, we construct $D_\attack$ by sampling from $D_\aux$ at random.
We leave open a question of whether more sophisticated methods
of constructing $D_\attack$ can lead to better attacks.

\paragraph{Shadow models and attack meta-classifier.}
The attacker relies on shadow models
in order to determine whether $\mathcal{F}$ is generated
from $f$ trained on a dataset with property $p$ or not.
To this end, the attacker trains $n$ ``shadow" models
that resemble $f$.
In particular, it generates
 training datasets $D_\shadow^i$, half of them exhibiting the property and half not, labeled as~$p$
and~$\bar{p}$ accordingly. These datasets could be obtained by resampling from~$D_\aux$.
Each shadow model $f_{\shadow}^i$ is trained on a dataset $D_\shadow^i\cup D_\adv$ using the same way as the target central model~$f$.
Once $f^i_{\shadow}$ is trained, the attacker queries it using $D_\attack$ and combines
inference results to form a feature vector~$\mathcal{F}_i$ associated with~$p$ or~$\bar{p}$, depending on its training data.

After training all shadow models, the adversary has a set of features $\mathcal{F}_i$
with the corresponding property label $p_i\in \{p, \bar{p}\}$.
The adversary then trains a meta-classifier on the pairs $\{(\mathcal{F}_i,p_i)\}_i$ using any binary
classification algorithm.
For example, logistic regression is sufficient for attacks in our experimental evaluation.

The attacker carries out its attack as follows.
Once the target model $f$ is trained on the joined data of the attacker and \victimW{} party,
the attacker queries the model using $D_\attack$
to obtains the feature representation of the target model, $\mathcal{F}$.
It then feeds $\mathcal{F}$ to its meta-classifier
and obtains a prediction for the sensitive property $p(\mathbf{a}_\victim)$.

\paragraph{Single-party attack.} We explained the attack strategy for the multi-party case since this is the primary focus of this work. However, we can easily adapt
the attack to the single-party case:
the only change that has to be made to the attack description above
is by setting $D_\adv$ to an empty set.
As highlighted in~Table~\ref{tbl:attackcmp}, besides being
the first attack on property leakage in the centralized multi-party setting,
our attack is also the first to show that dataset properties can be leaked
in the black-box setting.

\vspace{-5pt}
\paragraph{Fine-grained attack.}
The above attack shows how an adversary can learn whether some property is present
in a dataset or not. The attacker can extend this binary property attack and distinguish
between multiple properties $P = \{p^1, p^2, \ldots\}$.
It simply generates shadow training datasets for each property and then
trains a meta-classifier to predict one of the properties in $P$ based
on attack vector~$\mathcal{F}$.
For example, $P$ can be a set of possible ratios of females
to other values, and the attack meta-classifier will try to distinguish whether it is
10:90, 50:50 or 90:10 split.
In the experimental evaluation, we show that this attack is effective in learning fine-grained distribution
of sensitive attributes as well as identifying how the distribution of a sensitive attribute has changed
after the model was updated with new data.

\paragraph{Scope.}
\label{sec:scope}
\added{This work focuses on understanding the leakage of population-level properties of the training dataset.
Since our threat model is similar to that of the attacker who is able to infer individual record-level attributes~\cite{Song2020Overlearning,Congzheng,DBLP:conf/ccs/FredriksonJR15}, our setting allows for record-level leakage as well.
Albeit, the attack strategy needs to be changed in order to train shadow models that capture the difference between inputs with different attribute values.
Importantly, for both the record-level and population-level attribute inference attack, the attacker --- here and in~\cite{Song2020Overlearning,Congzheng} --- is assumed to know the domain of an attribute it is trying to infer (e.g., \texttt{Gender} taking values male, female, or other). Hence, similar to prior work~\cite{10.1145/3243734.3243834,DBLP:journals/corr/abs-1805-04049},
our attack cannot infer a sensitive attribute with a large, potentially unbounded, domain (e.g., such as \texttt{Name} for which the attacker may not be able to enumerate all possible values).}


\section{Experimental Setup}
\label{sec:evaluation}

The goal of our experiments is to evaluate the efficacy of the attack in Section~\ref{sec:attack}
to learn population-level
properties about a sensitive attribute in the multi-party and single-party machine learning setting.
We then aim to understand how the difference in machine learning models (e.g., logistic regression and neural network models),
dataset type (e.g., tabular data, text or graph),
access to the model through its weights or inference interface,
and attribute correlation influence attack accuracy.

\vspace{-5pt}
\subsection{Benchmark Datasets}
\vspace{-5pt}
We evaluate our attack on five datasets described below.
The datasets, sensitive attributes, machine learning model tasks, and the type of correlations between the sensitive attribute, other attributes, and the final task are summarized in Table~\ref{correlation_table}. 

\vspace{-5pt}
\paragraph{Health~\cite{health}} The Health dataset  (Heritage Health Prize) contains medical records of over $55~000$ patients.
Similar to the winners of the Kaggle competition,
we use $141$ features with \texttt{MemberID} and \texttt{Year} removed.
We group the \texttt{DaysInHospital} attribute
into two classes. The task, $Y$, is to predict if a patient will be discharged, $\texttt{DaysInHospital}=0$, or will stay in the hospital, $\texttt{DaysInHospital}>0$.
We consider two sensitive attributes to perform our attack on learning their distribution in the dataset of the benign party:
\texttt{Gender} and the number of medical claims \texttt{ClaimsTruncated}.

\vspace{-5pt}
\paragraph{Adult~\cite{kohavi1996scaling,lichman2013uci}}
The Adult dataset contains US census information including race, gender, income, and education level. The training dataset contains $32~561$ records with
$14$ attributes. We group the education level into four classes: `\texttt{Low}',  `\texttt{Medium-Low}', `\texttt{Medium-High}', `\texttt{High}'. \added{We use $12$ features with \texttt{Education} and \texttt{Fnlwgt} removed.}
The task is to predict the class of the \texttt{EducationLevel} (i.e., variable $Y$ for this dataset). 
We again consider two sensitive features whose distribution the attacker is trying to infer: \texttt{Gender} and \texttt{Income}. 

\vspace{-5pt}
\paragraph{Communities and Crime~\cite{lichman2013uci}}
The Communities and Crime dataset contains $1~994$ records with $122$ features relevant to per capita violent crime rates in the United States,
which was also used for evaluating fairness with respect to protected variables~\cite{pmlr-v97-creager19a}.
We remove the attributes that have missing data, resulting in $100$ attributes. The classification task is to predict the crime rate, i.e., the $Y$ variable is \texttt{CrimesPerPop}.
We group the crime rate into {three} classes based on ranges: `$<0.15$', `$[0.15,0.5]$' and `$>0.5$',
and the task is the multi-class prediction for the crime rate.
We consider total percentage of divorce \texttt{TotalPctDiv} and \texttt{Income} as sensitive features.

\vspace{-5pt}
\paragraph{Yelp-Health~\cite{yelp}}
The Yelp dataset contains 5 million reviews of businesses tagged with numerical ratings (1-5) and attributes such as business type and location.
We extract a healthcare-related subset that has $2~384$ reviews for pediatricians and $1~731$ reviews for ophthalmologists.
The classification task is to predict whether the review is positive (rating $>3$) or negative (rating $\le 3$).
The attack aims to predict the dominant value of the doctor \texttt{Specialty} of the benign party.

\vspace{-5pt}
\paragraph{Amazon~\cite{leskovec2007dynamics,amazon}}
The Amazon product co-purchasing network dataset contains product metadata and reviews information about $548~552$ different products such as books and music CDs.
For each product, the following information is available: the similar products that get co-purchased, product type, and product reviews.
We use a subset of $20~000$ products and construct a product co-purchasing network, where each node represents a product and the edge represents if there is at least one reviewer who rated both products, indicating that products are bought by the same user~\cite{doi:10.1080/15427951.2009.10129177}.
Each node is associated with one of 4 product types and an average review score from 0 to 5, including half-score reviews (i.e., 11 possible scores in total).
The classification task (for a recommendation system) is to predict the average review score of the node given the co-purchasing network and the product types.
Depending on the classification task, we split reviewer scores into 2 classes: positive vs. negative review, 6 classes: rounded integer review between 0,1.., 5 and 11 classes: the original review score.
The attack aims to predict whether the dominant value of the attribute \texttt{ProductType} of the benign party
is ``books''.

\subsection{Evaluation Methodology}\label{sec:eva}


\paragraph{\bf Target model $f$.} We train different target models depending on the dataset type. For tabular data, i.e., Adult, Health, and Crime, we train multinomial logistic regression and
fully-connected multi-layer perceptron neural networks (MLP). For the Adult and Crime datasets, we use an MLP network with one hidden layer of size 12 and the last layer with 4 and 3 output classes, respectively.
For the Health dataset, we use an MLP network with one hidden layer of size 20 and binary output.
{In later sections, a neural network model for tabular datasets always refers to an MLP network.}
In training our target models, we use the Adam \cite{kingma2014adam} optimizer, ReLu as the activation function, a learning rate of $0.01$, and a weight decay of $0.0001$. For the Yelp-Health dataset, we use the pre-trained glove embedding of dimension 50, a bidirectional LSTM layer of dimension 50. We then use one hidden layer of size 50 and dropout regularization with parameter 0.1 between the last hidden layer and the binary output. For the Amazon dataset, we train the target model using the Graph Convolutional Networks (GCN)~\cite{kipf2016semi} with 1 hidden layer of 16 units, Adam as the optimizer, ReLu as the activation function, a learning rate of $0.01$, and a weight decay of $0.0005$.
Each experiment is repeated 100 times, and all attack accuracies
are averaged over these runs.
\mpcrev{As noted in Section~\ref{sec:prelims}, our attacks are oblivious to how $f$ is trained, hence,
in the experiments training is done in the clear.}

\paragraph{\bf Dataset split.}
In the multi-party setting,
we consider two parties that contribute data for training the target model
where one of the parties is trying to learn information about the data of the other party.
For Adult and Health datasets, each party contributes 2 000 samples.
We use 10 000 \added{or 4 000} samples as $D_\aux$ to train the shadow models
and the attacker uses 1 000 samples in~$D_\attack$ to query the model
and obtain the attack vector for the meta-classifier. 
Table \ref{split_table} summarizes the splits for all other datasets.
In Section~\ref{sec:param} we show that a small number of samples
in $D_\attack$ can lead to high attack accuracy as well (e.g., 200 vs.~1 000 for the Amazon dataset).

\label{sec:dataatt_split}
\added{The distribution of the values of the sensitive attribute~$A$ in datasets is determined as follows.
We consider the default split of 33:67 in the attacker's data $D_\adv$ (e.g., 33\% of records are books). The attack is evaluated against several $D_\victim$ datasets for each possible split.
For example, we evaluate our attack on 100 $D_\victim$ datasets: half with 33:67 split and half with 67:33 split in Sections~\ref{sec:mps} and \ref{sec:sps}. Throughout all experiments, the $D_\aux$ always has 50:50 split. }

\begin{table}[t]
	\centering
	\resizebox{\linewidth}{!}
	{
		\begin{tabular}{@{}l | c  r  r  @{}}
			\toprule
			Datasets   & $\#D_\adv$, $\#D_\victim$ & \#$D_\aux$ &\#$D_\attack$   \\ 
			\midrule \midrule

			Health~\cite{health} & 2 000 & 10 000 \added{/ 4000}  &   1 000         \\  \midrule
			
			Adult~\cite{kohavi1996scaling,lichman2013uci} & 2 000 & 10 000 \added{/4000}   &   1 000       \\ \midrule			
			
			Crime~\cite{lichman2013uci} &    200 &    1 500 \added{/ 400} &         94        \\  		\midrule			
			
			Yelp-Health~\cite{yelp} & 1 000 & 1 200 & 200 \\ \midrule
			Amazon~\cite{leskovec2007dynamics} & 5 000 & 10 000 & 1 000
			\\
			
			\bottomrule
			
	\end{tabular}}
	\caption{Dataset split during the attack where $\#D_\attack$ is the number of inference queries the attacker makes to the model.}
	\vspace{-15pt}
	\label{split_table}
\end{table}

\vspace{-8pt}
\paragraph{\bf Attack setting.}
We report our main results on attack
in the black-box setting\added{; white-box results are deferred to~Appendix~\ref{sec:whiteboxapp}}\removed{ and report results on the white-box setting in Section~\ref{sec:whiteboxmain}}.
We use two different meta-classifiers depending on the target model.
For multinomial logistic regression, LSTM and GCN,
the meta-classifier model is a binary logistic regression model.
For MLP as the target model, we use a two-layer network with $20$ and $8$ hidden units and a learning rate of $0.001$.
The meta-classifier models are trained using Adam optimizer.

We perform the attack when the model is trained with the sensitive variable ($A$) and without it ($\bar{A}$).
For the $\bar{A}$ setting, the attribute $A$ is omitted from the machine learning pipeline, including the shadow model training
and construction of $D_\attack$. This setting allows us to understand the risk of leaking a sensitive attribute, even when that attribute is censored during training. 
For Yelp-Health, we report only $\bar{A}$ results as LSTM takes the text data, and $A$ would be an additional feature.


\begin{table*}[t]
	\centering
	{
		\begin{tabular}{@{}l | l|  l|  l  @{}}
			\toprule
			Datasets   & Sensitive attribute $A$ & Task $Y$ & Correlation   \\ 
              \midrule \midrule

			\multirow{2}{*} {Health~\cite{health}} &\texttt{Gender} & \multirow{2}{*}{\texttt{DaysInHospital}}   &                     \multirow{2}{*}{  $X \sim A, Y \bot A$}       \\ 
			                                         &      \texttt{ClaimsTruncated}   & &                           \\ \midrule
			
			\multirow{2}{*} {Adult~\cite{kohavi1996scaling,lichman2013uci}} &\texttt{Gender} & \multirow{2}{*}{\texttt{EducationLevel}}   &                   $X \sim A, Y \bot A$      \\ 
                                          &      \texttt{Income}   & &  $X \sim A, Y \sim A$                         \\ \midrule			
			
			\multirow{2}{*} {Crime~\cite{lichman2013uci}} &\texttt{TotalPctDivorce}& \multirow{2}{*}{\texttt{CrimesPerPop}}  &               \multirow{2}{*}{    $X \sim A, Y \sim A$}        \\ 
                                          &      \texttt{Income}   & &                        \\ 
\midrule			
			
 Yelp-Health~\cite{yelp} & \texttt{Specialty} & \texttt{ReviewRating} & $X \sim A, Y \bot A$\\ \midrule
 Amazon~\cite{leskovec2007dynamics} & \texttt{ProductType} & \texttt{ReviewScore} & $X \sim A, Y \sim A$
			\\
			
			\bottomrule
			
	\end{tabular}}
	\caption{Datasets, tasks and attribute-label correlation where $\sim$ and $\bot$ indicate {correlation} and no correlation, respectively.}
	\label{correlation_table}
\end{table*}

\vspace{-8pt}
\paragraph{\bf Types of experiments.} 
We study how correlations between attributes affect the attack. We show that information is leaked even
when $A$ is not correlated with the final task. 
We demonstrate our attack on attribute correlation as present in real dataset distributions (shown in Table~\ref{correlation_table}) as well as artificially injected correlation using a synthetic sensitive variable.
The latter allows us to control the correlation between the variables.

\noindent {{\em Real Data.}} For the experiments where all features are from the real data, including the sensitive variable, 
we set different variables as sensitive ($A$) for each dataset  and perform a black-box attack using a default split of 
33:67 for the sensitive attribute in the attacker's data ($D_\adv$). 

We compute the pairwise correlation among all the variables using Pearson correlation coefficient~\cite{pearson1895vii} for numerical-numerical variables, Cramer's V~\cite{cramer1999mathematical} for categorical-categorical variables, \removed{and ANOVA for categorical-numerical variables. } \added{point-biserial correlation coefficient~\cite{sheskin2020handbook} for binary categorical-numerical variables, and ANOVA for multi-level categorical-numerical variables.}
Based on the observed correlations, for each dataset, we identify the case among those introduced in~Section~\ref{sec:datamodel}.
Most scenarios correspond to $X \sim A, Y \sim A$. \added{Details on correlation factors for all datasets are deferred to Appendix~\ref{sec:correlation}.}
\removed{For Adult with $A$ as $\texttt{Gender}$,  Health, and Yelp-Health dataset, we observe that the correlation corresponds to $X \sim A, Y \bot A$. For Yelp-Health dataset, the scenario corresponds $X\sim A, Y\bot A$, because the {ANOVA p-value is $0.57$  for testing correlation between $Y$ and $A$} {point-biserial correlation coefficient between $Y$ and $A$ is 0.009}, and the review text is clearly correlated with the doctor specialty as shown in Table 4 in~\cite{DBLP:journals/corr/abs-1805-04049}.
For the Amazon dataset, the ANOVA p-value for testing correlation between $Y$ and $A$ is $7.6e-83$. We conjecture that the co-purchasing graph $X$ is also correlated with the product type $A$, and thus, the scenario corresponds $X\sim A, Y\bot A$.}

\noindent {{\em Synthetic Data.}} For synthetic experiments, we create a new synthetic attribute as our sensitive variable~$\SensA$ for the Adult and Health datasets.
We add a correlation of $\SynthA$ to a subset of variables in the dataset, denoted as $\subX\subseteq X$, and the target variable~$Y$, depending on the cases outlined in Section~\ref{sec:datamodel}.
\added{We introduce the correlation by replacing attribute values in $X'$ and/or $Y$ for each record with values that have an injected correlation with $A$.}
For Adult dataset, $\subX$ is \texttt{Income}, for Health dataset, $\subX = \{\texttt{DrugCountAve}, \texttt{LabCountAve},
\texttt{ClaimsTruncated}\}$. The variable $\SynthA$ takes values $< 5$ or $>5$ that are split using 33:67 ratio
in the {adversarial party's dataset. The \victimW{} party has two possible splits: 33:67 ratio and 67:33 ratio.}  The attacker's goal is to guess the distribution of $\SynthA$ in the data of $P_\victim$.
\label{sec:synthgen}


\vspace{-5pt}
\section{Attack Results}
\label{sec:experiments}
\vspace{-5pt}
We evaluate for attribute leakage in the following
settings:
the single-party case where an attacker learns the distribution of
an attribute in the training set and
the multi-party case where an attacker learns the distribution of
an attribute in the data of the honest party.
Apart from inferring the dominant attribute (e.g., there are more females than males in a dataset), we perform a fine-grained attack that learns a precise distribution of the two attribute values (e.g., 70\% of the dataset are females). We further use this fine-grained attack to infer the change in the attribute distribution in a model update scenario where the model is updated either due to a new party joining or new data arriving.
\randrev{Attack accuracy higher than the probability of a random correct guess is considered successful
as this indicates that confidential property (i.e., information about $P_{\victimW{}}$'s data)
will be leaked to the attacker in the majority of cases.}

We report our attack results in the stronger black-box setting for real, synthetic, and fine-grained experiments. We evaluate the white-box attack, where the attacker has access to model parameters, only on the synthetic data.
We summarize our key findings below:
\begin{itemize}
\squish
	\item Leakage of sensitive dataset properties in \victimW{} party's data is possible even when the sensitive attribute itself is {\em dropped} during training {and has low or no correlation with the final task.}
	We show that the attack accuracy drops only by a few percent when $A$ is not present in many cases.
	\item An adversary can learn the attribute properties of the \victimW{} party's data  irrespective of whether it contributes data  (multi-party)
	or not (single-party) to the training dataset.
	\item For the models and datasets considered in this paper, our property leakage attack is dataset and model-agnostic and works on tabular, text, or graph data.
	\item Fine-grained attacks can be used to predict a precise distribution of the attribute as well as learn the change in data distribution during model updates.
\end{itemize}

\begin{table*}[t]
\centering
\resizebox{0.8\linewidth}{!}
{
\begin{tabular}{@{}l |    c |  c  c |  l | l @{}}
\toprule
   \multirow{2}{*}{\begin{tabular}[c]{@{}l@{}} Datasets \\ (Output  Classes) \end{tabular}}   & {Model Type} & \multicolumn{2}{c|}{Attack Accuracy} & \multirow{2}{*}{$A$ } & \multirow{2}{*}{\added{\#} $\subX$}  \\ \cmidrule(r){3-4}
              &           & $A$            & $\bar A$           &                                          &                                    \\ \midrule \midrule

\multirow{2}{*} {\begin{tabular}[c]{@{}l@{}} Health (2) \end{tabular}}     & \multirow{2}{*} {Multi-layer Perceptron}             &    \removed{ .73} \added{.61}           &  \removed{.69} \added{.59}           &     {\begin{tabular}[c]{@{}l@{}}  \texttt{Gender}\end{tabular}}                &           \removed{15} \added{24}/139                            \\ \cmidrule{3-6}
 &                &     \removed{. 65} \added{.75}         &   \removed{. 63}\added{.71}                                          &     {\begin{tabular}[c]{@{}l@{}} \texttt{ClaimsTruncated}\end{tabular}}                &           \removed{66} \added{54}/139                         \\ \midrule
\multirow{2}{*}{\begin{tabular}[c]{@{}l@{}} Adult (4) \end{tabular}}   &      \multirow{2}{*} {Logistic Regression}   & \removed{.75} \added{.83}           &  \removed{.77} \added{.81}                  &       {\begin{tabular}[c]{@{}l@{}} \texttt{Gender} \end{tabular}}        &            \removed{9} \added{5}/11                        \\  \cmidrule{3-6}

    &          &    .98          &   .96                          &       {\begin{tabular}[c]{@{}l@{}} \texttt{Income} \end{tabular}}        &            \removed{7} \added{9}/11                        \\ \midrule
\multirow{2}{*}{\begin{tabular}[c]{@{}l@{}}Crime (3)\end{tabular}}    &          \multirow{2}{*} {Multi-layer Perceptron}                 & .61               &  .59                                        &           {\begin{tabular}[c]{@{}l@{}} \texttt{TotalPctDivorce} \end{tabular}}              &      \removed{13} \added{26}/98                  \\  \cmidrule{3-6}

   &                    &    .78            &     .60                                     &      {\begin{tabular}[c]{@{}l@{}} \texttt{Income}  \end{tabular}}   &     \removed{33} \added{38}/98      \\  \midrule
  Yelp-Health (2)  &       LSTM                &   -           & \removed{ .70} \added{.74}                                          &       {\begin{tabular}[c]{@{}l@{}} \texttt{Specialty} \end{tabular}}        &           {review text}                    \\  \midrule
   Amazon (2)    &   GCN                     &         .86           &  .72                                          &     {\begin{tabular}[c]{@{}l@{}} \texttt{ProductType} \end{tabular}}        &           {graph}                   \\  \midrule
     Amazon (6)   &    GCN                           &        .62          &  .63                            &      {\begin{tabular}[c]{@{}l@{}} \texttt{ProductType} \end{tabular}}        &           {graph}                     \\  \midrule
      Amazon (11)            &  GCN         &         .67          &  .61                                          &      {\begin{tabular}[c]{@{}l@{}} \texttt{ProductType} \end{tabular}}        &           {graph}                     \\ 
   
    \bottomrule

\end{tabular}}
\vspace{0.1cm}
	\caption{Multi-Party Setting: Black-box attack accuracy for predicting the value of the distribution of sensitive variable~$A$ in the dataset of $P_\victim$.
	The attacker tries to guess whether values of $A$ are split as 33:67 or 67:33 in $D_\victim$ when its own data $D_\adv$ has 33:67 split.
	Columns $A$ and $\bar{A}$ report the accuracy when the sensitive variable is used for training and not, respectively. 
	$X'$ indicates with which features in the dataset and with how many of them $A$ is correlated.
	Since attack accuracy based on a random guess is $0.5$, the attacker is always successful in determining the correct distribution.}
	\label{tbl:realdata-multi} 
	\vspace{-5pt}
\end{table*}

\vspace{-5pt}
\subsection{Multi-Party Setting}\label{sec:mps}

\textbf{Real Data.}
Table~\ref{tbl:realdata-multi} shows the attack accuracy for correlations observed in the real distribution of datasets, \added{with the larger size of $D_\aux$ as listed in Table \ref{tbl:attackcmp}. The attack accuracy with the smaller size of $D_\aux$ is deferred to Table \ref{tbl:realdata-smallaux} in Appendix~\ref{sec:adtable}.}
We see that the attack accuracy is always better than a random guess in all experiments,
regardless of whether the sensitive attribute is included in the training data or not.


We make the following observations.
The attack accuracy for Adult data with \texttt{Income} as the sensitive attribute is the highest with 98\% and 96\% when the target model is trained with and without A, respectively. 
Overall, the attack accuracy ranges between 61-98\% when trained with sensitive variable ($A$) and 
59-96\% without ($\bar{A}$), respectively. The results for~$\bar{A}$ are always lower than with~$A$ but are, however, above the random guess baseline of 50\%.
For the Amazon dataset, we observe that attack accuracy is higher for fewer output classes. We confirm this observation later in Figure~\ref{fig:graph_removetrue}.
\added{We also note that the attack accuracy decreases as the size of $D_\aux$ decreases as shown in Appendix~\ref{sec:adtable}.}

To understand how the correlation between $A$ and other features influences the attack,
we determine which attributes $X' \subseteq X$ are correlated with $A$.
We set $\subX$ to variables based on their correlation factors. \removed{For example,
for the Health dataset, all features are binarized, and we use Cramer's V as the correlation factor
where the score between \texttt{DaysInHospital} and \texttt{Gender} is 0.09,
and thus, we deem them as uncorrelated.
Among the rest of the attributes, we identify 15 attributes
that have Cramer's V scores with \texttt{Gender} greater than 0.15
and, hence, assign them to~$\subX$. }\added{Details on how $X'$ of each dataset was determined based on correlation factors
is deferred to Appendix~\ref{sec:correlation}.}
\added{In Table \ref{tbl:realdata-multi}, \# $X'$ denotes the number of attributes correlated with the sensitive attribute $A$. We note that simultaneously controlling the number of correlated attributes and their correlation strength is hard on real data, so we also use synthetic datasets.}
\added{We observe that, for the same dataset, the attack accuracy increases with
a higher number of correlated attributes $X'$ and the sensitive attribute $A$.}\


\added{We show the accuracies for both the pooled model and the \victimW{}  party's local model in Table \ref{tbl:utility} in Appendix~\ref{sec:adtable}.} {Across all these experiments, we observe a utility increase ranging from \removed{ 0.5\% to 4.5\%} \added{ 0.58\% and 5.90\%} for the \victimW{} party, which motivates the \victimW{} party to collaborate and train a joint target model with the other party.
}

%

\textbf{Synthetic Data.}
Table~\ref{tbl:synthetic-multi} shows our results with a synthetic variable $\SynthA$ introduced in the Adult and Health dataset for the multi-party setting.
Here, we train the same dataset using both logistic regression and the neural network model (MLP).
Recall that the synthetic attribute is introduced to imitate a sensitive variable to control its correlation with other variables.
To this end, we create datasets for different correlation criteria among the sensitive variable $A$, the output $Y$, and the remaining variables $X$.
We report two findings.

First, logistic regression models appear to be at a higher risk, with average attack accuracy being higher as compared to neural network models:
{$84.5\%$ vs.~$71.3\%$ for Adult and $80.2\%$ vs.~$70.8\%$ for Health datasets}.
We suspect that this is mainly due to their simple architecture, which is easy to learn using a meta-classifier. 

Second, the attack works well (greater than 74\%) when the sensitive variable $A$ is correlated with the target variable $Y$ irrespective of its relation with $X$,
i.e., cases where $Y\sim A$.
The attack accuracy is almost equal to
a random guess when $Y\bot A$.
Recall that in the case of $X\sim A$, not all features used for training are correlated with $A$ but only those
in a subset of $X$, $\subX$.
To understand this scenario further, we reduced the number
of features used during training to 3 (we refer to this setting as \RedactT{} in the tables).
As the number of training features decreases, the correlation signal between $A$ and $\subX$ becomes stronger, and the logistic regression model can capture that.
\removed{However, as we showed earlier, in realistic datasets, most of the variables are highly correlated with the sensitive variable and hence exhibit better attack accuracy.}


Our experiments for the case when both $X$ and $Y$ are independent of the sensitive variable $A$ exhibit attack accuracy that is close to a random guess. This is expected as the variable has no correlation that the model can memorize, and hence we exclude them from Table~\ref{tbl:synthetic-multi}.

\vspace{-5pt}
\subsection{Single-Party Setting} \label{sec:sps}
In addition to our motivating scenario of the multi-party setting, we evaluate the efficacy of our attack in the single-party setting where the attacker does not contribute towards the training data.
For example, this corresponds to a scenario where a model is trained on data from only one hospital and is offered as an inference service for other hospitals. 
Table \ref{tbl:synthetic-single} shows the result for our attack using synthetic data for the Adult and Health dataset when the model is trained using both logistic regression and neural networks.
We see that the attack in the single party setting is stronger since
the adversary does not provide its own data, which may dilute the signal from the other party. For the case where $Y\sim A$, the attack accuracy is higher than 90\%, even \added{if} the attribute itself is not used during training. This shows that our attack is highly successful even when the attacker does not participate in the training process.

\begin{table}[t]
\centering
\resizebox{\linewidth}{!}
{
\begin{tabular}{@{}l | rr   rr | rr rr @{}}
\toprule
Model & \multicolumn{4}{c|}{Logistic Regression}  &  \multicolumn{4}{c}{Neural Network} \\ \midrule
Datasets & \multicolumn{2}{c}{Adult}                                                           & \multicolumn{2}{c|}{  Health}                                                                 & \multicolumn{2}{c}{Adult}                                                           & \multicolumn{2}{c}{  Health}                                                                  \\ \midrule 
             Synthetic Variable                      & $A$                & $\bar{A}$                & $A$                &  $\bar{A}$                      & $A$                & $\bar{A}$                & $A$                &  $\bar{A}$                       \\  \midrule
 $X \sim A, Y \sim A$                 &    1.00            &        1.00       &   1.00            &        1.00      & .90           & .84        & .79           & .95                      \\ 
 $X \bot A, Y \sim A$            &  1.00              &  1.00           &              .99          &       1.00         & .98           & .98      & .74           & .98                     \\ 

$X \sim A, Y \bot A$                  &  .65              &   .57                     &    .52           &       .41     & .52           & .52    & .52           & .51                       \\  
 $X \sim A, Y \bot A$ (\RedactT{})                 & .79               & .75                &     .78         &      .72       & .51           & .45   &   .54     &    .63         \\  
\bottomrule
\end{tabular}}	\vspace{0.2cm}
	\caption{Multi-party setting: Black-box attack accuracy for predicting whether the values of (sensitive) synthetic variable~$A$ in the data of the honest party
		are predominantly $\textless 5$ or $\textgreater 5$.
		The attack accuracy is evaluated on 100 $D_\victim$ datasets: half with 33:67 and half with 67:33 split.
		A synthetic correlation with $A$ is added to the variables $X$ and $Y$ depending on the specific case.
		\RedactT{}~corresponds to the setting where only 3 attributes are used for training instead of all data.
		Attack accuracy based on a random guess is $0.5$.}
	\label{tbl:synthetic-multi} 
\end{table}

\begin{table}[t]
	\centering
	\resizebox{\linewidth}{!}
	{
		\begin{tabular}{@{}l | rr   rr | rr rr @{}}
			\toprule
			Model & \multicolumn{4}{c|}{Logistic Regression}  &  \multicolumn{4}{c}{Neural Network} \\ \midrule
			Datasets & \multicolumn{2}{c}{Adult}                                                           & \multicolumn{2}{c|}{  Health}                                                                 & \multicolumn{2}{c}{Adult}                                                           & \multicolumn{2}{c}{  Health}                                                                  \\ \midrule 
			Synthetic Variable                      & $A$                & $\bar{A}$                & $A$                &  $\bar{A}$                      & $A$                & $\bar{A}$                & $A$                &  $\bar{A}$                       \\  \midrule
			$X \sim A, Y \sim A$                 &    1.00            &        1.00       &  \removed{ .979}\added{.98}            &        1.00      & .98           & .99        & .92           & .95                       \\ 
			$X \bot A, Y \sim A$            &  1.00              &  1.00           &             \removed{ .978}\added{.98}          &       1.00         & .99           & 1.00     & .89           & .98                     \\ 
			
			$X \sim A, Y \bot A$                  & \removed{ .666}\added{.67}              &   \removed{.595}\added{.60}                     &   \removed{ .481}\added{.48}           &     \removed{ .529}\added{.53}    & .56           & .52    & .52           & .49                       \\  
			$X \sim A, Y \bot A$ (\RedactT{})                 &\removed{ .856}\added{.86}               & \removed{.744 }\added{.74}               &    \removed{ .608}\added{.61}         &     \removed{ .620}\added{.62}      & .68           & .66   &   .54     &    .61        \\  
			\bottomrule
	\end{tabular}}	\vspace{0.2cm}
	\caption{Single-party setting: Black-box attack accuracy with synthetic data.}
	\label{tbl:synthetic-single} 
\vspace{-5pt}
\end{table}

\subsection{Fine-grained Attack}
\label{sec:finegrained}
Information leaked about attribute values can be either in terms of a binary signal,
i.e., which attribute value is dominant
in the dataset or an exact distribution. The results above show the leakage of the former.
To learn information about the exact distribution, we present a variation of our main attack called the fine-grained attack.
For this attack, we train a 5-class meta-classifier model
that outputs whether a particular value of the sensitive attribute
appears in 10\%, 30\%, 50\%, 70\%, or 90\% of the dataset. Note that we train only one meta-classifier model with 5 output classes, but the attacker can perform a more systematic binary search over the distribution by training multiple meta-classifier
models.
We apply this attack in two settings.

\vspace{-5pt}
\paragraph{Leakage of Attribute Distribution.}

We evaluate on the Adult dataset using a synthetic variable $\SynthA$ as well as the gender variable.
Table \ref{tbl:fine-grained} shows the results for our fine-grained attack for predicting the precise distribution of the sensitive variable. 
The row $30:70$ corresponds to the setting where 30\% of records in $D_\victim$ have the value of the sensitive attribute $\SynthA$ less than 5.
Here, the attacker tries to guess the split of $30:70$ among five possible splits of $10:90$, $30:70$, etc. The baseline accuracy is~20\% because the attacker wishes to distinguish between 5 splits.
Since the attack accuracy is always higher than the random guess,
the attacker can successfully find the correct ratio
by training a meta-classifier that distinguishes between different splits of the sensitive attribute values.
Similar to the observation in Section \ref{sec:mps},
we observe that logistic regression has higher attack accuracy than neural networks. 
The attack accuracy for the real data with \texttt{gender} as the sensitive attribute is consistently greater than the $20\%$ baseline for random guessing for all the distributions.

\begin{table}[t]
\centering
\resizebox{\linewidth}{!}
{
\begin{tabular}{@{} c | cc | cc| c @{}}
\toprule
\multicolumn{1}{c|}{{\begin{tabular}[c|]{@{}c@{}}Distribution \\ of $\SynthA$ in $D_\victim$: \end{tabular}}} &
 \multicolumn{2}{c |}{\begin{tabular}[c|]{@{}l@{}}LR \\ Synthetic $A$ \end{tabular}} &
  \multicolumn{2}{c |}{\begin{tabular}[c|]{@{}l@{}}NN \\Synthetic  $A$   \end{tabular}} &
  \multicolumn{1}{c }{\begin{tabular}[c]{@{}l@{}}LR \\ $A$: \texttt{Gender} \end{tabular}} \\ 
                            & $A$            & $\bar{A}$            & $A$            &  $\bar{A}$    &  $\bar{A}$      \\ \midrule
 10 : 90                                                                                                                            &     .994 &  .998       	& .84	&                               .89&.44  \\
 30 : 70                                                                                                                     &  .993  & .991                  	& .79	&                             .79 &.59 \\
 50 : 50                                                                                                                                & .999 &  .997            & .79	&                                 .73 &.50\\
 70 : 30                                                                                                                       &.997  &.989              &	.73		&                               .71 &.46\\
 90 : 10                                                                                                                         & .993 &  .998               &	.72 &                               .77& .53\\ 
\bottomrule
\end{tabular}}
	\caption{Fine-grained attack accuracy for predicting the precise distribution of sensitive variable $A$ in $D_\victim$ in the synthetic setting $X \bot A, Y \sim A$, and real data setting when $A$ is \texttt{Gender} on the Adult dataset.
		Attack accuracy based on a random guess is $0.2$.}
\label{tbl:fine-grained}
\vspace{-25pt}
\end{table}

\paragraph{Model Update Setting.}
\label{sec:snapshot}
We apply the fine-grained attack to learn the change in the distribution of an attribute value given access to an updated version of a model.
In this attack, the malicious party initially obtains a model that is jointly trained on $D_\mathsf{honest1}$ and $D_\adv$. Later, another honest party $D_ \mathsf{honest2}$ joins, and a new model is trained on the three parties' data.
The attacker tries to infer the dominant value of the sensitive attribute of $P_\mathsf{honest2}$ given the \added{original and the} updated model.
\removed{The attacker applies the fine-grained attack on the updated model as follows.}
\added{\label{sec:cmtrevB6}It uses a fine-grained attack against both models, as result learning
a dominant value in $D_\mathsf{honest1}$ and $D_\mathsf{honest1} \cup D_\mathsf{honest2}$.
It then compares the two and infers how $D_\mathsf{honest2}$ has affected the distribution.
If the split is dominated by the same attribute value
in both models, the attacker uses this attribute value distribution as its guess.
Otherwise, the attacker makes a guess that the other attribute value is dominated in $D_\mathsf{honest2}$.}
\removed{If the predicted split is still dominated by the same class as in the attack on the original model,
it suggests that the dominant value in $D_ \mathsf{honest2}$ is the same as in $D_\mathsf{honest1}$.}
Table \ref{tbl:snapshot} shows the results for our attack in the model update setting using synthetic data for the Adult dataset. The attack accuracy is almost close to 100\% for the synthetic case and ranges from 63\% to 86\% for the \texttt{Gender} variable which is higher than a random guess of 50\%.

\begin{table}[t]
	\centering
	\resizebox{\linewidth}{!}
	{
		\begin{tabular}{@{} c | c | c |c @{}}
			\toprule
			\begin{tabular}[c|]{@{}c@{}}Distribution \\ of $\SynthA$ in $D_\mathsf{honest1}$: \end{tabular}&
			\begin{tabular}[c|]{@{}l@{}}Distribution \\ of $\SynthA$ in $D_\mathsf{honest2}$:  \end{tabular}&
			\begin{tabular}[c]{@{}l@{}}LR \\ Synthetic $A$  \end{tabular} &
			 \begin{tabular}[c]{@{}l@{}}LR \\ $A$: \texttt{Gender} \end{tabular}\\ \midrule
		
            \multirow{2}{*}{30:70} & 30:70 & 1.00 &.87\\
                                           & 70:30 & .99 &.72\\ \midrule
           \multirow{2}{*}{70:30} & 30:70 & .99&.63\\
               & 70:30 & 1.00&.85\\               
                         
            \bottomrule
	\end{tabular}}
	\caption{Model update setting: attack accuracy for predicting the dominant value of sensitive variable $A$ in $D_\mathsf{honest2}$ in the synthetic setting $X \bot A, Y \sim A$ and real data setting when $A$ is \texttt{Gender} on Adult dataset when $A$ is removed from the training data. $D_\adv$ has 50:50 split. Attack accuracy based on a random guess is $0.5$.}
	\label{tbl:snapshot}
\end{table}

\begin{figure}[t]
\centering
\includegraphics[scale=0.45]{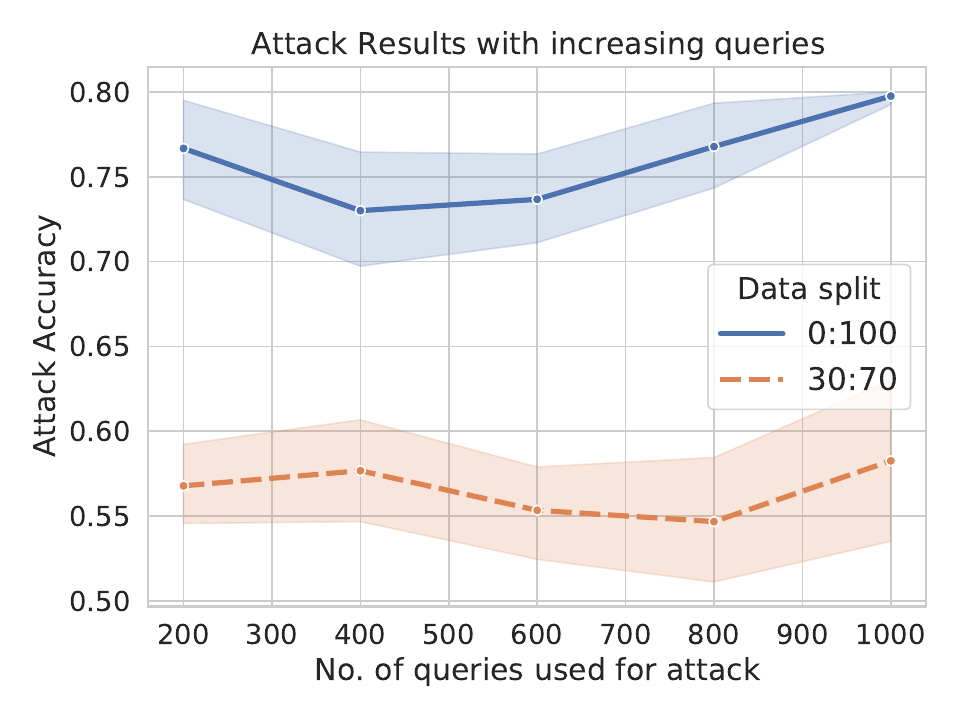}
\vspace{-5pt}
\caption{Attack accuracy for leaking sensitive attribute \texttt{ProductType} on the Amazon graph data (11 output classes) as the number of queries to the model increases.}
\label{fig:graph_queries}
\vspace{-5pt}
\end{figure}

\subsection{Attack Parameters}
\label{sec:param}
We perform ablation experiments to understand the effect of varying the number of queries, distribution of the sensitive attribute and the number of output classes on the attack accuracy.
We use the Amazon graph data for these experiments where, as before, \texttt{ProductType} is the sensitive attribute, and \texttt{ReviewScore} is the target.


\vspace{-11pt}
\paragraph{Number of queries.}\vspace{-2pt}
We compute the attack accuracy for two different splits of values of the sensitive attribute, 0:100 (all books) and 30:70 (70\% books, 30\% of other products), and train the model to predict one of 11 review scores \added{averaged over 10 runs}.
\removed{Figure~\ref{fig:graph_queries} shows that increasing the number of queries increases the attack accuracy, more rapidly for 0:100 split as compared to 30:70 split.} \added{Figure~\ref{fig:graph_queries} shows the effect of increasing the number of queries on the attack accuracy}. Note that the number of queries also correspond to the input features of our attacker classifier. \removed{We observe that for the 0:100 split, as few as 200 queries are sufficient to get an attack accuracy above 75\%.}\added{We observe that changing queries does not significantly impact the attack accuracy. With 1000 queries, attack accuracy is up to 80\% for the 0:100 split and $\approx$59\% for 30:70 split. }

\begin{figure}[t]
\centering
\vspace{-5pt}
\includegraphics[scale=0.45]{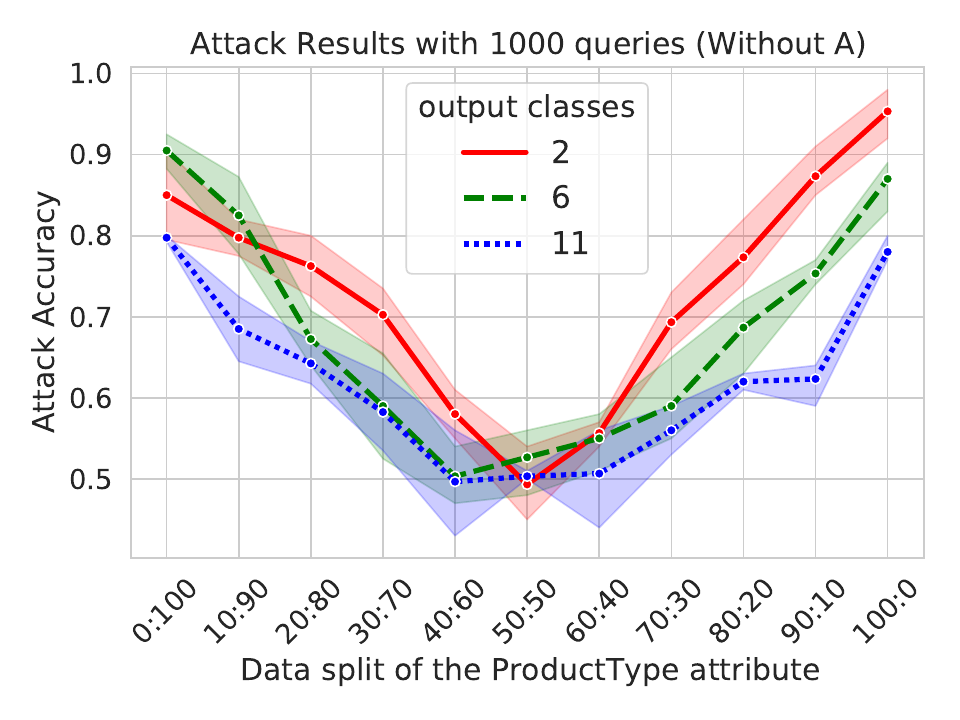}
\vspace{-5pt}
\caption{Attack accuracy for the Amazon graph data when the sensitive attribute \texttt{ProductType} is not used during training for different numbers of output classes across different distributions (splits).}
\label{fig:graph_removetrue}
\end{figure}

\vspace{-5pt}
\paragraph{Attribute distribution and number of output classes.}
Figure~\ref{fig:graph_removetrue} shows the results for the GCN trained on the Amazon dataset for 2, 6 and 11 output classes for the review score.
\added{We evaluate for all the splits between 0:100 to 100:0.
First, we observe that the attack accuracy drops as the ratio of the sensitive attribute values changes from 0:100 to 50:50 and increases again gradually from 50:50 to 100:0.
This is because our primary attack is designed to identify the dominant attribute value. For inferring the distribution in the balanced range, the attacker can perform our fine-grained attack discussed in Section~\ref{sec:finegrained}.}
 \removed{However, note that the attack accuracy is always above the random guess baseline of 50\%.} Next, we observe that the attack accuracy is lower for a higher number of output classes such as 6 and 11 as compared to 2. \added{This could be due to lower
 number of input features that are given to the attack classifier when there are lower number of output classes --- the classifier is able to learn the attribute distribution better when the information is divided among fewer features thus resulting in a lower dimension input.}
\added{Similar trends are observed in Figure~\ref{fig:graph_removefalse} in Appendix when $A$ is used during training.}

 \removed{For output class 2, attack accuracy for a model trained using the sensitive attribute (Figure~\ref{fig:graph_removetrue}) mostly does not change and drops less gradually than other classes in Figure~\ref{fig:graph_removetrue}.}

\ifWithRevisions
\subsection{White-box Attack}
\fi
\label{sec:whiteboxmain}
\removed{
We also performed experiments where the attacker has access to the model parameters,
i.e., the white box setting.
As for the meta-classifier model, we use
a two-layer network with $200$ and $50$ hidden units and learning rate $0.001$.
Each meta-classifier is trained based on 100 shadow models using Adam optimizer.
Here, the meta-classifier takes as input model parameters as opposed to model inferences.
Table~\ref{tbl:white-box} shows the results.
For logistic regression, the results are similar to those in Table~\ref{tbl:synthetic-multi} for the black-box setting.
However, the attack accuracy for neural networks (MLP) reduces significantly.
This was also noted in the work by~\cite{10.1145/3243734.3243834}.
One reason for this is that it is hard for a naive meta-classifier to learn the structure of equivalent symmetrical weights of neural networks.
Indeed, one of the contributions of~\cite{10.1145/3243734.3243834} is a technique for identifying this symmetry.}

\section{Defenses}
\label{sec:defences}

In the previous section, we saw that removing
the sensitive attribute from the dataset is not an effective solution
due to the correlations that exist between the attributes.
Disentangling data representation through
variational-auto-encoders~\cite{Higgins2017betaVAELB,pmlr-v28-zemel13,pmlr-v97-creager19a} allows one
to obtain mutually independent variables
for representing the data. Intuitively,
the removal of this variable before decoding
the record for further down-stream tasks would
lead to better censorship.
Similarly, adversarial learning has also been proposed for
learning a privacy-preserving data filter in a multi-party setting~\cite{Hamm2015PreservingPO}
and a privacy-preserving record representation~\cite{9aa5ba8a091248d597ff7cf0173da151}.
Unfortunately, such techniques do not have provable
worst-case guarantees and have been shown ineffective in the
privacy context~\cite{Song2020Overlearning}.

Differential privacy~\cite{privacybook} guarantees
record-level privacy, that is,
whether a particular record is in their dataset or not.
However, differential privacy does not protect population-level
properties of a dataset~\cite{privacybook,10.1145/2020408.2020598}.
In fact, a differentially private algorithm with high utility aims to learn population properties without sacrificing individual privacy.
Group differential privacy is an extension of differential privacy
that considers the privacy of a group of~$k$ correlated records\added{, as a result
one way of achieving it is to increase, for example, Laplace noise, proportional to~$k$.}
\removed{where privacy
guarantee drops linearly with the size of the group.}
Though it can be applied to preserve the privacy of all records in each party's dataset by setting $k$ to the size of each party's data,
\removed{this would lead to high utility sacrifices.}
\added{\label{cmt:revB11}depending on the setting, it can effect utility as
even with $k=1$ accuracy of models have been shown to drop~\cite{abadi,DBLP:conf/kdd/SongS19}.}

In settings with more than two parties, where the attacker controls
only one party, the signal weakens as it is harder for the adversary
to identify the mapping between a property and a party whose
data exhibits it. This was also noted by~Melis~\textit{et al.}~\cite{DBLP:journals/corr/abs-1805-04049} in the federated learning setting
with a small number of parties.


\section{Related work}
\label{sec:related}
Membership attacks on machine learning models
aim to determine whether a certain record was part of a training dataset or not~\cite{DBLP:conf/sp/ShokriSSS17,Salem:NDSS19}.
These attacks train shadow models that are similar to the target model
and use their output (e.g., posterior probabilities over all classes) to build a meta-classifier that classifies records as members of the training
data or not based on inference results of the target model on the record in question.
A recent link stealing attack on graphs can be seen as a type of a membership attack that tries to infer
whether two nodes have a link between them in the training graph~\cite{He2020StealingLF}.

Attribute inference attacks~\cite{Song2020Overlearning,DBLP:conf/ccs/FredriksonJR15}, on the other hand,
aim to determine the value of a sensitive attribute
for a single record.
For example, the authors of~\cite{Song2020Overlearning} study leakage of a sensitive value from
a latent representation of a record in the model (i.e., a feature extractor layer);
an attacker can obtain such intermediate record representations from having access to model parameters.
They show that an attribute of a record, even if censored using adversarial learning, can be leaked.
Hitaj~\textit{et al}~\cite{10.1145/3133956.3134012} show that a malicious party
can construct class representatives from a model trained in federated learning setting.

The work by Ganju~\textit{et al.}~\cite{10.1145/3243734.3243834}
and Ateniese~\textit{et al.}~\cite{10.1504/IJSN.2015.071829} are closest to ours as they also consider leakage of \emph{dataset properties}.
Different from this work, their attack is set in a single-party setting and requires a \emph{white-box access} to the model,
i.e., its parameters, that may not always be possible (e.g., when the model access is via cloud-hosted interface).
Since the number of model parameters in neural networks can be very large (several million),
approaches that are based on sophisticated methods for reducing network representation
are required~\cite{10.1145/3243734.3243834}.
We show that attacks based on a combination of inferences
and logistic regression as a meta-classifier are sufficient to learn attribute distribution.
%
%

Property leakage in a multi-party learning has been
demonstrated only
in federated setting~\cite{DBLP:journals/corr/abs-1805-04049}.
In this setting an attacker obtains a gradient computed on
a small batch of records (e.g., 32) and tries to learn how a sensitive feature is distributed
in the batch.
This setting is arguably easier from the attacker point of view:
an attacker gains access to a much more granular computation on the data
compared to the access to a query interface of the final model trained on the whole dataset, as considered in this paper.
Moreover, previous work on dataset property leakage~\cite{10.1145/3243734.3243834,DBLP:journals/corr/abs-1805-04049,10.1504/IJSN.2015.071829} did not consider the case
when the sensitive attribute is {\em removed} from the data and the effect it has on the success of their attacks.

Recently, Zanella-B{\'e}guelin~\etal~\cite{brockschmidt2019analyzing}
have demonstrated leakage of text and general trends in the data used to update next word prediction model.
Salem~\etal~\cite{DBLP:conf/uss/0001B0F020}, on the other hand, consider granular leakage about records used to update the model:
record labels and their features.
{Similar to our work, Salem~\etal use a probing dataset to query the models to obtain the posterior difference.}
This output is then given to an encoder-decoder framework to reconstruct the meaning
of the difference between posteriors of the initial and updated models.
Our model update attack, in comparison, is about identifying the distribution
of a sensitive feature in the dataset used to update the model
and requires a simple machine learning architecture.

\section{Conclusion}

We demonstrate an attack, set in the centralized multi-party machine learning, that lets one of the parties learn sensitive properties about \removed{an}other part\removed{y's}\added{ies'} data\removed{sets}.
The attack requires only black-box access to the model and can extract the distribution of a sensitive attribute with small number of
inference queries.
We show that trivial defenses such as excluding a sensitive attribute from training are insufficient to prevent leakage.
Our attack works on models for tabular, text, and graph data
and datasets that exhibit various correlation relationships among attributes and class labels. Finally, we note that existing techniques for secure computation and differential privacy
are \added{either} not directly applicable to protect leakage of population-level properties \added{or do so at a high cost}.

\section*{Acknowledgements}We thank Marcella Hastings and the anonymous reviewers for their valuable comments on the paper.

\bibliographystyle{abbrv}



%

\bibliography{main,olyarefs}

\appendix

\appendix

\section{Attribute Correlation in Datasets}
\label{sec:correlation}

This section provides information on  correlation cases  for the datasets and attributes in~Table~\ref{correlation_table}.

\textbf{Health Dataset.}  We measure the correlations between \texttt{Gender} or \texttt{ClaimsTruncated} and the $133$ categorical attributes and $6$ numerical attributes by Cramer's V scores and point biserial correlation coefficients, respectively. With \texttt{Gender} as the sensitive attribute, we identify $22$ categorical attributes that have Cramer's V scores greater than $0.15$ and $2$ numerical attributes that have point biserial correlation (absolute value) greater than $0.1$. The attributes that have the highest Cramer's V are \texttt{sp10} (0.218), \texttt{noSpecialities} (0.212),  \texttt{noProviders} (0.208), \texttt{noVendors} (0.201). To give a overview of correlations including weak correlation with other attributes, we identify $17$ attributes that have Cramer's V scores within the range $[0.1, 0.15]$ and $37$ attributes Cramer's V scores within the range $[0.5, 0.1]$. The Cramer's V score between \text{DaysInHospital} and \texttt{Gender} is 0.09, and thus, we deem them as uncorrelated.
With \texttt{ClaimsTruncated} as the sensitive attribute, we identify $50$ categorical attributes (e.g., \texttt{sp1} (0.42), \texttt{sp2} (0.51), \texttt{pcg1} (0.41), etc.) that have Cramer's V scores greater than $0.15$, and $4$ numerical attributes that have point biserial correlation (absolute value) greater than $0.1$. The score between \texttt{DaysInHospital} and \texttt{ClaimsTruncated}  is 0.13, and we set them  uncorrelated.

\textbf{Adult Dataset.} 
We measure the correlations between \texttt{Gender} or \texttt{Income} and the $7$ categorical attributes and $4$ numerical attributes by Cramer's V scores and point biserial correlation coefficients, respectively. We list all the correlation factors in Table \ref{adult_table}, as $X$ only has $11$ attributes. For \texttt{Gender}, we identify $4$ categorical attributes that have Cramer's V scores above $0.15$ and $1$ numerical attribute that has point biserial correlation coefficients above $0.1$. The sensitive attribute \texttt{Income} has a high Cramer's V score with $5$ categorical attributes and the target variable \texttt{EducationLevel}, as well as high point biserial correlation coefficients with $4$ numerical attributes.

\begin{table}[t]
	\centering
	\resizebox{0.7\linewidth}{!}
	{
		\begin{tabular}{@{}l | c  c  @{}}
			\toprule
			Attributes   & \texttt{Gender} & \texttt{Income}   \\ 
			\midrule \midrule
			\multicolumn{3}{c}{Cramer's V scores}   \\  \midrule
			
			\texttt{EducationLevel} & 0.042 & 0.326       \\ \midrule			
			
			\texttt{MaritalStatus} &   0.466 &    0.448      \\  		\midrule			
			
			\texttt{Occupation} & 0.435 & 0.329  \\    \midrule
			
			\texttt{Relationship} & 0.650 & 0.454  \\    \midrule
			
			\texttt{Race} & 0.119 & 0.099  \\    \midrule
			
			\texttt{NativeCountry} & 0.059 & 0.096  \\    \midrule
			
			\texttt{Income} & 0.217 & -  \\    \midrule
			
			\texttt{Gender} & - & 0.217  \\    \midrule \midrule
			
			\multicolumn{3}{c}{point biserial correlation coefficients}   \\  \midrule
			
			\texttt{Age} & 0.082 & 0.229      \\  \midrule			

			\texttt{CapitalGain} & 0.049 & 0.221         \\  \midrule
		
			\texttt{CapitalLoss} & 0.047 & 0.150       \\  \midrule

			\texttt{HoursPerWeek} & 0.231 & 0.230     \\
			
			\bottomrule
			
	\end{tabular}}
	\caption{Correlation factors for the Adult dataset.}
	\label{adult_table}
\end{table}

\textbf{Crime Dataset.}
Since all features are numerical, we measure the Pearson correlation coefficients. Table \ref{crime_table} shows the number of attributes that have the coefficients within a certain range. We use $0.4$ as the threshold to determine $\subX$. The target variable \texttt{CrimesPerPop} is correlated with both \texttt{TotalPctDivorce} and \texttt{Income}, with correlation coefficients $0.553$ and $-0.424$, respectively.

\begin{table}[t]
	\centering
	\resizebox{0.7\linewidth}{!}
	{
		\begin{tabular}{@{}l | c  c  @{}}
			\toprule
			Range   & \texttt{TotalPctDivorce} & \texttt{Income}   \\ 
			\midrule \midrule
			
			$[0.5,1]$ & 15 &  34      \\ \midrule			
			
			$[0.4,0.5)$ &   11 &    4      \\  		\midrule			
			
			$[0.3,0.4)$ & 31  & 22  \\    \midrule
			
			$[0.2,0.3)$ & 4  & 12  \\    \midrule
			
			$[0.1,0.2)$ & 14 & 19  \\    

			\bottomrule
			
	\end{tabular}}
	\caption{Correlation factors for  Crime dataset.}
	\label{crime_table}
\end{table}

\textbf{Yelp-Health and Amazon Datasets.} For Yelp-Health dataset, the point biserial correlation coefficients between \texttt{Specialty} and \texttt{ReviewRating} is 0.009, hence,  the scenario corresponds $X\sim A, Y\bot A$. The review text is clearly correlated with the doctor specialty as  in Table 4 in~\cite{DBLP:journals/corr/abs-1805-04049}.
For the Amazon dataset, since the \texttt{ProductType} has 4 levels, we use the ANOVA to test whether the differences between the means of \texttt{ReviewScore} across different product types are statistically significant. The ANOVA p-value is $7.6e-83$. We conjecture that the co-purchasing graph $X$ is also correlated with the  \texttt{ProductType}.
 and hence $X\sim A, Y\sim A$.

\section{Additional Results}\label{sec:adtable}
\begin{figure}[t]
\centering
\includegraphics[scale=0.45]{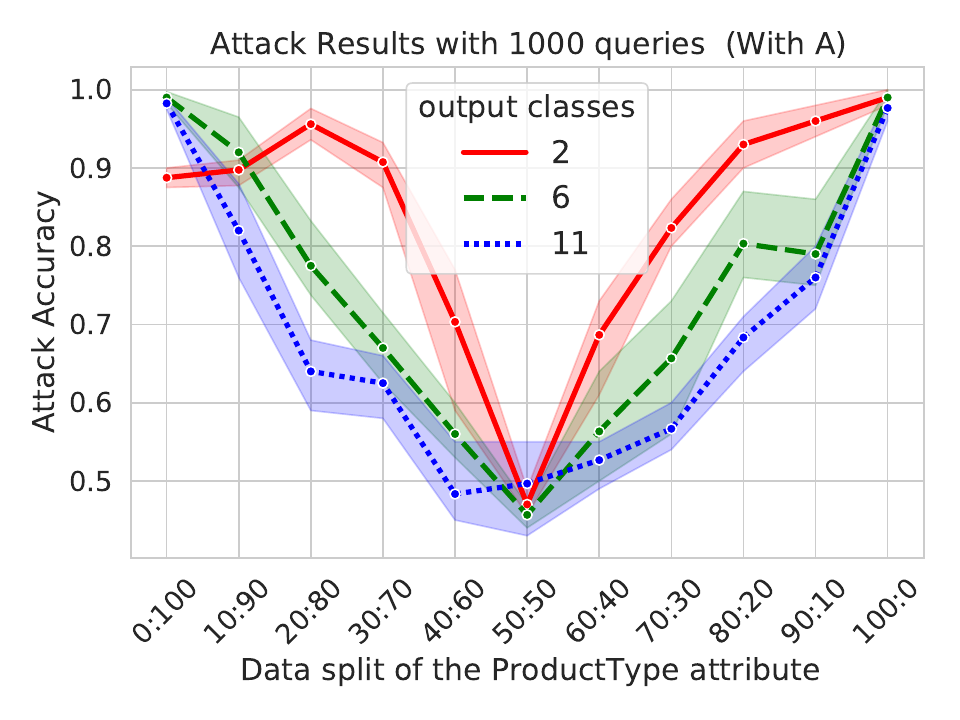}
\caption{Attack accuracy for the Amazon  data with the sensitive attribute \texttt{ProductType} used during training.}
\label{fig:graph_removefalse}
\vspace{-1cm}
\end{figure}

We present attack results for Health, Adult and Crime dataset with smaller size of $D_\aux$ from Table \ref{split_table}. We show accuracies for both pooled model and the \victimW{}  party's local model and the utility increase in Table \ref{tbl:utility}.
Figure~\ref{fig:graph_removefalse} complements results in Section~\ref{sec:param} on Amazon dataset trained with the sensitive attribute $A$.

\begin{table*}[t]
	\centering
	{
		\begin{tabular}{@{}l |    c |  c  c |  c  l @{}}
			\toprule
			\multirow{2}{*}{\begin{tabular}[c]{@{}l@{}} Datasets \\ (Output  Classes) \end{tabular}}    & \multirow{2}{*}{Sensitive Attribute} & \multirow{2}{*}{Pooled Accuracy } & \multirow{2}{*}{Local Accuracy}  & \multirow{2}{*}{Utility Increase} \\ 
			&           &             &            &                                          &                                    \\ \midrule \midrule
			
			\multirow{2}{*} {\begin{tabular}[c]{@{}l@{}} Health (2) \end{tabular}}                          &     {\begin{tabular}[c]{@{}l@{}}  \texttt{Gender}\end{tabular}}               &     85.22\% &     84.64\%  &     .58\%                   \\ \cmidrule{2-6}
			                                                     &     {\begin{tabular}[c]{@{}l@{}} \texttt{ClaimsTruncated}\end{tabular}}                &      76.63\% &    73.56\%   &       3.07\%                       \\ \midrule
			\multirow{2}{*}{\begin{tabular}[c]{@{}l@{}} Adult (4) \end{tabular}}                 &       {\begin{tabular}[c]{@{}l@{}} \texttt{Gender} \end{tabular}}    &       73.23\%  &           72.46\%  &      .76\%                \\  \cmidrule{2-6}

			                            &       {\begin{tabular}[c]{@{}l@{}} \texttt{Income} \end{tabular}}        &      71.14\% &    70.43\%  &      .71\%                                            \\ \midrule
			\multirow{2}{*}{\begin{tabular}[c]{@{}l@{}}Crime (3)\end{tabular}}                                                       &           {\begin{tabular}[c]{@{}l@{}} \texttt{TotalPctDivorce} \end{tabular}}              &  74.52\%  &      72.35\% &    2.17\%            \\  \cmidrule{2-6}
			
			                                                    &      {\begin{tabular}[c]{@{}l@{}} \texttt{Income}  \end{tabular}}   &  72.81\% &  71.30\%    &  1.51\%      \\  \midrule
			Yelp-Health (2)                                                        &       {\begin{tabular}[c]{@{}l@{}} \texttt{Specialty} \end{tabular}}        &   86.28\% & 80.38\% & 5.90\%                  \\  \midrule
			Amazon (2)                                                      &     {\begin{tabular}[c]{@{}l@{}} \texttt{ProductType} \end{tabular}}        &         76.80\% & 76.28\% & .62\%               \\  \midrule
			Amazon (6)                                             &      {\begin{tabular}[c]{@{}l@{}} \texttt{ProductType} \end{tabular}}        &         45.92\% & 42.50\% & 3.42\%                   \\  \midrule
			Amazon (11)                                                            &      {\begin{tabular}[c]{@{}l@{}} \texttt{ProductType} \end{tabular}}        &  27.94\%   &    26.09\%     &    1.85\%                     \\ 
			
			\bottomrule
			
	\end{tabular}}
\caption{Test accuracies of the model trained on pooled dataset and the model trained only on \victimW{} party's data. The split in the \victimW{} party is $33:67$ based on the sensitive attribute. }
\label{tbl:utility}
\end{table*}

\begin{table}[t]
	\centering
	\resizebox{\linewidth}{!}
	{
		\begin{tabular}{@{}l |   c   c|    c |  l | l @{}}
			\toprule
			\multirow{2}{*}{\begin{tabular}[c]{@{}l@{}} Datasets (Classes)  \\    {Model Type} \end{tabular}} & \multicolumn{2}{c|}{Attack Accuracy} & \multirow{2}{*}{$A$ } & \multirow{2}{*}{\# $\subX$}  \\ \cmidrule(r){2-3}
			&           & $A$            & $\bar A$           &                                          &                                    \\ \midrule \midrule
			
			\multirow{2}{*} {\begin{tabular}[c]{@{}l@{}} Health (2) \\ MLP \end{tabular}}               &     .59          &   .55              &     {\begin{tabular}[c]{@{}l@{}}  \texttt{Gender}\end{tabular}}                &           {24}/139                            \\ \cmidrule{2-6}
			               &     . 67          &   . 56                                          &     {\begin{tabular}[c]{@{}l@{}} \texttt{ClaimsTruncated}\end{tabular}}                &           {54}/139                         \\ \midrule
			\multirow{2}{*}{\begin{tabular}[c]{@{}l@{}} Adult (4) \\ LR \end{tabular}}     & .73           &  .76                  &       {\begin{tabular}[c]{@{}l@{}} \texttt{Gender} \end{tabular}}        &          {5}/11                        \\  \cmidrule{2-6}

			        &    .84          &   .91                          &       {\begin{tabular}[c]{@{}l@{}} \texttt{Income} \end{tabular}}        &            {9}/11                        \\ \midrule
			\multirow{2}{*}{\begin{tabular}[c]{@{}l@{}}Crime (3) \\ MLP \end{tabular}}                & .60               &  .56                                        &           {\begin{tabular}[c]{@{}l@{}} \texttt{TotalPctDivorce} \end{tabular}}              &     {26}/98                  \\  \cmidrule{2-6}
			
			                   &    .62            &     .60                                     &      {\begin{tabular}[c]{@{}l@{}} \texttt{Income}  \end{tabular}}   &     {38}/98      \\ 
			
			\bottomrule
			
	\end{tabular}}
	\caption{Multi-Party Setting: Black-box attack accuracy for predicting the value of the distribution of sensitive variable~$A$ in the dataset of $P_\victim$. We use smaller size of $D_\aux$ listed in Table \ref{split_table}, while all other settings are the same as in Table \ref{tbl:realdata-multi}.}
\label{tbl:realdata-smallaux} 
\end{table}

\begin{table}[t]
	\centering
	\resizebox{\linewidth}{!}
	{
		\begin{tabular}{@{}l | rr   rr | rr rr @{}}
			\toprule
			Model & \multicolumn{4}{c|}{Logistic Regression}  &  \multicolumn{4}{c}{Neural Network} \\ \midrule
			Datasets & \multicolumn{2}{c}{Adult}                                                           & \multicolumn{2}{c|}{  Health}                                                                 & \multicolumn{2}{c}{Adult}                                                           & \multicolumn{2}{c}{  Health}                                                                  \\ \midrule 
			Synthetic Variable                      & $A$                & $\bar{A}$                & $A$                &  $\bar{A}$                      & $A$                & $\bar{A}$                & $A$                &  $\bar{A}$                       \\  \midrule
			$X \sim A, Y \sim A$               & .90           & .94      &   .85          &               .97    & .54           & .49        & .65          & .61      \\ 
			$X \bot A, Y \sim A$             & .95           & .93           &                  .81        &                .80              & .57           & .53   & .63          & .56                \\ 
			
			$X \sim A, Y \bot A$                 & .54           & .53                   &        .50       &               .53       & .56           & .53      & .54          & .51           \\  
			$X \sim A, Y \bot A$ (\RedactT{})                & .75           & .63                &    .76         &              .68 &     .55           & .50     & .55          & .45     \\  
			\bottomrule
	\end{tabular}}

	\caption{White-box attack accuracy for predicting whether the values of sensitive variable $A$ in $D_\victim$, the data of the honest party,
		are predominantly $\textless 5$ or $\textgreater 5$.
		The attack accuracy is evaluated on 100 $D_\victim$ datasets: half with 33:67 and half with 67:33 split
		and $D_\adv$ has 33:67 split.
		A synthetic correlation with $A$ is added to the variables $X$ and $Y$ depending on the specific case.
		\RedactT{}~corresponds to the setting where only 3 attributes are used for training instead of all data.
		Attack accuracy based on a random guess is $0.5$.}
	\label{tbl:white-box} 

\end{table}

\textbf{White-box Attack Results}
\label{sec:whiteboxapp}
Additionally, we performed experiments where the attacker has access to the model parameters,
i.e., the white box setting.
As for the meta-classifier model, we use
a two-layer network with $200$ and $50$ hidden units and learning rate $0.001$.
Each meta-classifier is trained based on 100 shadow models using Adam optimizer.
Here, the meta-classifier takes as input model parameters as opposed to model inferences.
Table~\ref{tbl:white-box} shows the results.
For logistic regression, the results are similar to those in Table~\ref{tbl:synthetic-multi} for the black-box setting.
However, the attack accuracy for neural networks (MLP) reduces significantly.
This was  noted in the work by~\cite{10.1145/3243734.3243834}.
One reason is that it is hard for a naive meta-classifier to learn the structure of equivalent symmetrical weights of neural networks.
Indeed, one of their contributions  is a technique for identifying this symmetry.

\end{document}